\documentclass{article}
\usepackage{graphicx} 
\usepackage{subcaption}
\usepackage{verbatim}
\usepackage{amssymb}
\usepackage{multirow}
\usepackage{booktabs}
\usepackage{url}

\title{VCRScore: Image captioning metric based on V\&L Transformers, CLIP, and precision-recall}

\author{Guillermo Ruiz, Tania Ramírez, Daniela Moctezuma}
\date{February 2024}

\begin{document}
\maketitle
\begin{abstract}
    Image captioning has become an essential Vision \& Language research task. It is about predicting the most accurate caption given a specific image or video. The research community has achieved impressive results by continuously proposing new models and approaches to improve the overall model's performance. Nevertheless, despite increasing proposals, the performance metrics used to measure their advances have remained practically untouched through the years. A probe of that, nowadays metrics like BLEU, METEOR, CIDEr, and ROUGE are still very used, aside from more sophisticated metrics such as BertScore and ClipScore.
    Hence, it is essential to adjust how are measure the advances, limitations, and scopes of the new image captioning proposals, as well as to adapt new metrics to these new advanced image captioning approaches.
    This work proposes a new evaluation metric for the image captioning problem. To do that, first, it was generated a human-labeled dataset to assess to which degree the captions correlate with the image's content. Taking these human scores as ground truth, we propose a new metric, and compare it with several well-known metrics, from classical to newer ones. Outperformed results were also found, and interesting insights were presented and discussed.

\end{abstract}


\section{Introduction}
Nowadays, the Vision \& language research area is continuously growing, with tasks such as Image Captioning, Visual Question-Answering, and Image generation, among others.
This kind of task involves text and images, which turn it into a very challenging problem.

Particularly, the Image Captioning (IC) task has been studied recently, and it can be very useful in many diverse applications, such as content retrieval, intelligent video surveillance systems applications, computer-human interfaces, etc.
State-of-the-art proposals to deal with the IC task are rapidly updating, they went from being based on traditional approaches, based on probabilistic models, to the current Vision Transformers. Nevertheless, evaluation metrics have been almost untouched through the years. Proof of this even today the metrics based on $n$-grams matching are still very popular in the IC Task~\cite{moctezuma2023video, Cui_2018_CVPR,wang2019hierarchical,rahman2019chittron}.
Moreover, there are more lexical-based metrics such as METEOR, ROUGE, SPICE, and CIDEr.
Despite that new semantic-based metrics have been proposed in the last few years, such as BertScore, BLEURT, and CLIP, there is much space for improvement yet~\cite{GONZALEZCHAVEZ2024117071}. This improvement is related to doing better metrics to evaluate the Image Captioning task. We know that the best way to do that is human-based, manually tagging the system's output, but this is very unpractical and highly expensive in time and money, and it is also very exhausting, so it is very necessary to have good and updated performance metrics.
As in any machine learning problem, how the Image Captioning models are tested is crucial to know how much we have done, how much is necessary to do, what routes to take, and which to avoid.

\subsection{Problem statement}
\label{sec:probstatement}

In any Natural Language Processing (NLP) tasks, such as automatic translation, image captioning, etc., comparing a set of reference sentences $X$ and a prediction or candidate $\hat{X}$ is usually used. This comparison is made by a metric which is a function $f(X,\hat{X})\in \mathbb{R}$.
Theoretically, better metrics have a higher correlation with human scores.
The metrics can be organized into some categories
\cite{zhang2019bertscore}:

\begin{itemize}
    \item $n$-gram based (such as BLEU, METEOR, ROUGE)
    \item Edit-Distance-based (such as CIDER)
    \item Embedding-based (such as BertScore, ClipScore)
    \item Learned or trainable (such as Beer~\cite{stanojevic2014beer}, Blend~\cite{ma2017blend}, and also our proposal)
\end{itemize}

In this work, a learned metric is proposed that deals with and considers both aspects, semantic and lexical, as well as has a good correlation with humans.
Our proposal is learned because we tackle the problem as a regression model where our $y$ or dependent variable, is the score provided by humans in our dataset generation as well as another found in the literature.
Several experiments were done, trying to find the best metric.

\subsection{Contribution}

We proposed a new Image Captioning metric to evaluate the model's output in both senses, lexical and semantical. To achieve that, it was necessary to build a new dataset through human tagging using as input 100 images from the MSCOCO dataset, the output of five state-of-the-art models, and the sixth one from the original set of captions from MS-COCO corresponding to the selected image. This overall process is described in Section~\ref{sec:tagging}.

Our main contributions are:
\begin{itemize}
    \item A new human-tagging dataset, corresponding to 600 image-caption score values.
    \item An analysis of the results of the tagging process, as well as the taggers' agreement and a comparison between models.
    \item A metric proposal based on a model trained with data, trying to approximate the same performance as human taggers\footnote{available in https://github.com/ResearchInfoGeo/VCRScore}. 
    \item A deep evaluation and comparison with other most used evaluation metrics.
\end{itemize}

The paper is organized as follows, the section~\ref{sec:tagging} details the entire procedure done by generating our dataset and its analysis and evaluation as well as the model's performance evaluation done by humans. Section~\ref{sec:proposal} describes all the components related to our metric proposal. Experimental results and their analysis are described in Section~\ref{sec:expandana}, and finally, Section~\ref{sec:conclusions} concludes.

\section{Dataset generation}
\label{sec:tagging}

We generate a new dataset using a set of images from the MS-COCO dataset\cite{lin2014microsoft} and the caption prediction from five state-of-the-art models in addition to one caption related to the image from the MS-COCO dataset. Theoretically, this last caption is the best possible since humans did it. This produces a dataset with 600 image-captions evaluation pairs. The models used to generate these captions were diverse, from those proposed many years ago to those more recent, they are: BLIP2~\cite{li2023blip2, blip2}, OFA~\cite{wang2022ofa}, Convcap~\cite{AnejaConvImgCap17},  Meshed-Memory Transformer (M$^2$)~\cite{cornia2020m2}, and Show Attend And Tell (SAAT)~\cite{pmlr-v37-xuc15}. In addition, we used the fifth reference from the MS-COCO dataset as another input for the assessment.

The main idea is to build a metric that learns the relationship or score provided by humans when they evaluate more and less adequate image-caption pairs.

\subsection{Tagging process}
\label{tagginganalysis}

As mentioned, a new dataset was generated from the knowledge of four taggers evaluating the captions from 100 images and the captions predictions from five different models. Furthermore, humans generated a correct caption from the MS-COCO dataset. That means a total of 600 pairs of image-caption to be evaluated by humans.
To do that, we built a simple tool to show the images and a randomly selected caption. 

\begin{figure}[!h]
    \centering
    \includegraphics[scale=0.3]{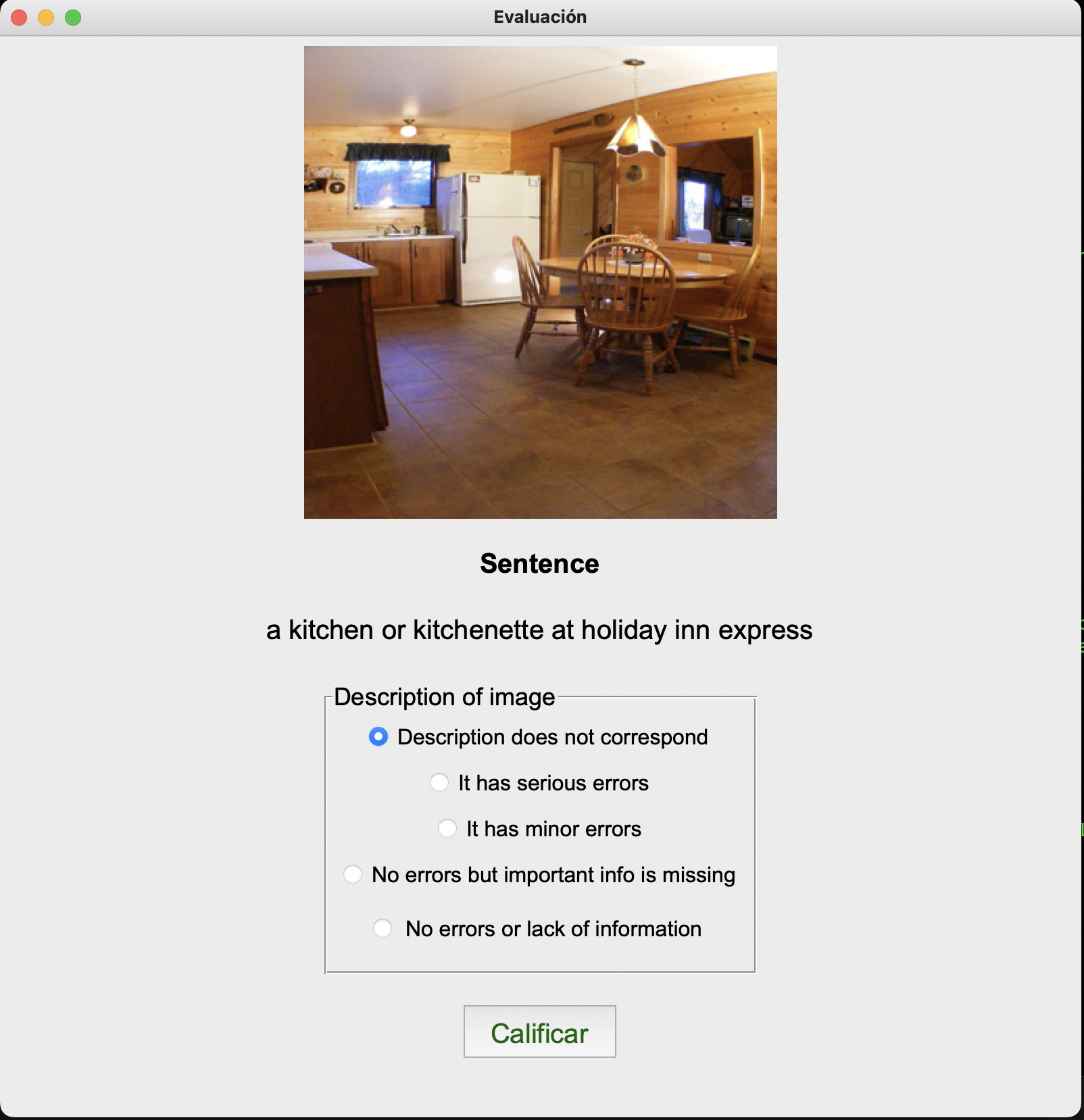}
    \caption{Screenshot of the simple tagging designed tool}
    \label{fig:apptania}
\end{figure}

A screenshot of this tool can be seen in Figure~\ref{fig:apptania} where it can be seen that there is an image displayed and also a caption related to it. Also, there are five score options to be selected by the human tagger, they are:

\begin{enumerate}
    \item No error or lack of information (best possible option, value 1),
    \item No error, but important info is missing (second best possible option, value 0.75)
    \item It has minor errors (third best possible option, value 0.5)
    \item It has serious errors (second worst possible option, value 0.25)
    \item Description does not correspond (worst possible option, value 0)
\end{enumerate}

Human taggers must select one that reflects the score of this image-caption relationship.

Four human taggers did this process. Each tagger meticulously evaluated 600 image-caption pairs, comprehensively assessing the association on a rank scale of 0, 0.25, 0.5, 0.75, and 1. This thorough evaluation process ensures a comprehensive understanding of the image-caption association.

This scheme was selected according to a caption-quality work \cite{levinboim2019quality}, where the authors declare the importance of helpfulness and correctness in an image pair caption evaluated by a tagger. The above helps to diagnose a model's quality or similarity to an image description done by a human. Helpfulness measures information about the image, and correctness is the correspondence between the sentence and the image; they used three scales for each characteristic.
For the measurement of predisposition or agreement, the Likert scale is more popular used in social science research; the primary purpose is to have a scale to estimate the strongly agree to strongly disagree \cite{likert1932technique}, generally used to two or eleven points, but the widely employed is the five-points, each scale designated to a numeric value \cite{abu2020qos}. Based on that, we decided to use a five-point scale because it provides more variety than just three levels.
Another consideration is the interval between re-tagging to minimize the memory effects; the authors of \cite{otter1995relation} recommend at least six weeks.

In this work, we merged both ideas, generated five scales as previously mentioned, and the whole tagging procedure was done twice with a time gap of three months.
So, in the end, we have a set of 600 scores produced twice by the same tagger, meaning 600 evaluations in two different periods.

\subsection{Taggers performance analysis}
After obtaining these data, we evaluate the tagger's agreement to get the best possible and highest-quality score for each image-caption pair.
We know that tagger agreement is significant in measuring the quality of the generated data.
That means reliability is key in selecting which data will be trusted to represent the phenomena of interest rather than fake or spurious data.
Assessing data quality and measuring agreement between human experts is a popular topic, as in~\cite{wiebe1999development}, where agreement patterns are analyzed to find disagreements related to bias among experts, then corrections can be made automatically.
As we know, reliability is the highest source of validation of assessment data. Low reliability tells us about high discordance or inconsistent data~\cite{downing2004reliability}.
Two agreement measurement scores were used, $\alpha$ of Krippendorf~\cite{krippendorff2009testing}, and the $\tau$ of Kendall~\cite{Puka2011,kendall1948rank}.

The Krippendorff’s $\alpha$~\cite{krippendorff2009testing} is a quantitative score that accounts for chance agreement, allowing for any number of observers and missing data, and also being suitable for many data types such as nominal, ordinal, interval, etc. $\alpha = 1$ means perfect agreement, while $\alpha = 0$ means that the observed agreement cannot be differentiated from chance. Negative $\alpha$ values indicate disagreement, \textit{i.e.} taggers somehow ``agree in how to disagree.''
$\alpha$, is a ratio score between observed $D_o$ and expected disagreement $D_e$.
The $\alpha$ computation is as follows:
\begin{equation}
     \alpha =  1 - { {D_o}\over{D_e}}
\end{equation}
where $D_o$  is the observed disagreement and $D_e$ is the disagreement expected by chance. For more technical information about $\alpha$ calculation, please refer to \cite{krippendorff2011agreement}. In our experiments, we used the Library Krippendorff implemented in Python\footnote{https://pypi.org/project/krippendorff/}.

We also decided to use an additional inter-rater reliability assessment metric, the $\tau$ of Kendall. Kendall's $\tau$ measures the degree of correlation or strength of the relation between two sets of values (X's, Y's).

The $\tau$ of Kendall is calculated as follows:
\begin{equation}
    \tau = {C - D \over {m}}
\end{equation}
where $m$ is the total possible comparisons from $n$ pairs of observations (X and Y).
$m$ is calculated:
\begin{equation}
    m = {n(n-1) \over {2}}
\end{equation}
So, $C$ is the number of concordant pairs, and $D$ is the opposite case. Concordant refers if a difference of a pair of points ($X_i$, $Y_i$) and ($X_j$, $Y_j$) have the same sign, that means if the sign of $X_i - Y_i$ is the same as $X_j - Y_j$. We used the implementation of the Scipy library from Python implementation\footnote{https://docs.scipy.org/doc/scipy/reference/generated/scipy.stats.kendalltau.html}.
For more details about Kendall's $\tau$, please revise the corresponding references~\cite{Puka2011,kendall1948rank}.

As we have four evaluations in each period, that means, 8 data evaluations in total, we proceed to calculate the agreement between the taggers.
We calculated both, the $\alpha$ of Krippendorf and the $\tau$ of Kendall between the human-generated scores.
Table~\ref{tab:agreemen} shows the agreement values of our four taggers compared to their first evaluation stage and the second one.

\begin{table}
    \centering
    \caption{Agreement score between first and second evaluation stage per Tagger}
    \label{tab:agreemen}
    \begin{tabular}{|c|c|c|}\hline
    \textbf{Tagger} & \textbf{$\alpha$ of Krippendorf}  &  \textbf{$\tau$ of Kendall}\\ \hline
       Tagger1  & 0.83  &0.73 \\ \hline
       Tagger2  & 0.48  & 0.46\\ \hline
       Tagger3  & 0.76 &0.64 \\ \hline
       Tagger4  & 0.79  &0.71 \\ \hline
    \end{tabular}

\end{table}

The table shows that the Tagger2 is more atypical than the rest because their agreement is low in both the $\alpha$ of Krippendorf and the $\tau$ of Kendall, which can be interpreted as Tagger2 differed more from the first and second evaluation periods.
Therefore, we did an extra analysis and put together all the evaluations from the two stages.
Table~\ref{tab:poretapas} shows different results from several data combination measurements. The agreement of the four evaluations in the first stage is 0.62 of Krippendorf $\alpha$, the same, but from the second stage is $0.60$, which is a very similar result. Furthermore, we analyzed what would happen if the data of one tagger were eliminated, and we see that the highest Krippendorf $\alpha$ value was 0.71 when we removed data from Tagger2. This analysis was not done with Kendall because it uses only data pairs, and we have a set of four data evaluations.

\begin{table}[!h]
    \centering
    \begin{tabular}{|c|c|c|}\hline
    \textbf{Combination} & \textbf{$\alpha$ of Krippendorf}  \\ \hline
       All taggers first stage & 0.62   \\ \hline
       All taggers second stage  & 0.60  \\ \hline
       All taggers two stages & \textbf{0.64}  \\ \hline
       All except tagger1, two stages & 0.62 \\ \hline
       All except tagger2, two stages &  \textbf{0.71}\\ \hline
       All except tagger3, two stages & 0.63 \\ \hline
       All except tagger4, two stages  & 0.61   \\ \hline

    \end{tabular}
    \caption{Agreement score in several data combinations}
    \label{tab:poretapas}
\end{table}

Now, we analyze the agreement between the data of all the taggers in each stage, which means considering the scores provided for all the taggers.

Table~\ref{tab:agreementtaggers} shows the different agreements calculated; it can be seen that the agreement is similar in all the cases, which means the agreement of the taggers in stage 1 was 0.62 and in stage 2 was 0.60, then despite Tagger2 seeming to be more inconsistent.
In conclusion, we used the data from all four taggers in their two stages because we considered that more data is better than a slight decrease in removing tagger2 from our dataset.

\begin{table}[!h]
    \centering
    \begin{tabular}{|c|c|c|}\hline
      \textbf{Time} & \textbf{$\alpha$ of Krippendorf}  &  \textbf{$\tau$ of Kendall}\\ \hline
       Phase 1  & 0.62 & \\ \hline
       Phase 2  & 0.60 & \\ \hline
       Phases 1 and 2 & \textbf{0.64} & \textbf{0.63}  \\ \hline
    \end{tabular}
    \caption{Agreement score of the evaluation in phases 1 and 2, from all the taggers.}
    \label{tab:agreementtaggers}
\end{table}

At the end of this process, we generated a unique value per each image-caption pair.
Determining the final score for each pair of caption-image samples is important. In this case, one can consider a simple voting scheme where the score predominant is the final assigned. Also, it can use the mean value from the scores provided per the caption-image pair.

We decided, at first, to use the two ways, the voting and mean of all the scores. Voting is calculated by considering the score that was most voted; we choose randomly in case of a tie.
In the case of the mean calculation, if we have a set of scores of image-caption, for instance, if we have the following set of scores $\{0.25, 0.50, 0.25, 0.50, 0.50, 0.50, 0.25, 0.50\}$, we calculate the mean of these values to get a unique score per image-caption pair, in this example, the final score is $0.4062$.

Both ways have positive and negative aspects; for instance, the voting scheme respects the majority opinion, but when there is a tie between two very different values, the resulting score can be slightly confusing. We think all the scores are considered in the mean calculation, but this provides a higher range of values than initially proposed.

\subsection{Models's performance}
\label{modelsperformance}

As mentioned above, we used five prediction models to be evaluated by the taggers. In this sense, one can calculate how good or bad all these prediction captions were done. Additionally, it is was evaluated one caption from the MS-COCO dataset, which was human-generated.
Considering that higher values correspond to better scores for the caption-image pair, we sum up all the scores per each of the 100 different images across all the prediction models and the MS-COCO caption. The results are in Table~\ref{tab:ranks} where both calculations are shown, the voting and mean approaches. As expected, the reference from the MS-COCO dataset was best evaluated with 93\% and 88.68\% on voting and mean, respectively. The second best evaluated, also expected, was the OFA model. These results are very similar to what can be found in the state-of-the-art.

\begin{table}[!h]
    \centering
    \begin{tabular}{|c|c|c|} \hline
    \textbf{Model}   & \textbf{Voting} & \textbf{Mean}  \\ \hline
    Humans  & \textbf{93.0} & \textbf{88.68}  \\ \hline
    OFA  & 89.0 & 86.5  \\ \hline
    BLIP2  & 84.5 & 82.25  \\ \hline
    M$^2$  & 64.0 & 65.62  \\ \hline
    Convcap  & 47.5 & 53.43  \\ \hline
    SAAT   & 26.0 & 33.34  \\ \hline

    \end{tabular}
    \caption{Evaluations for all the caption-image pairs}
    \label{tab:ranks}
\end{table}

\begin{figure}[!h]
\centering
\begin{subfigure}{0.45\textwidth}
    \includegraphics[width=\textwidth]{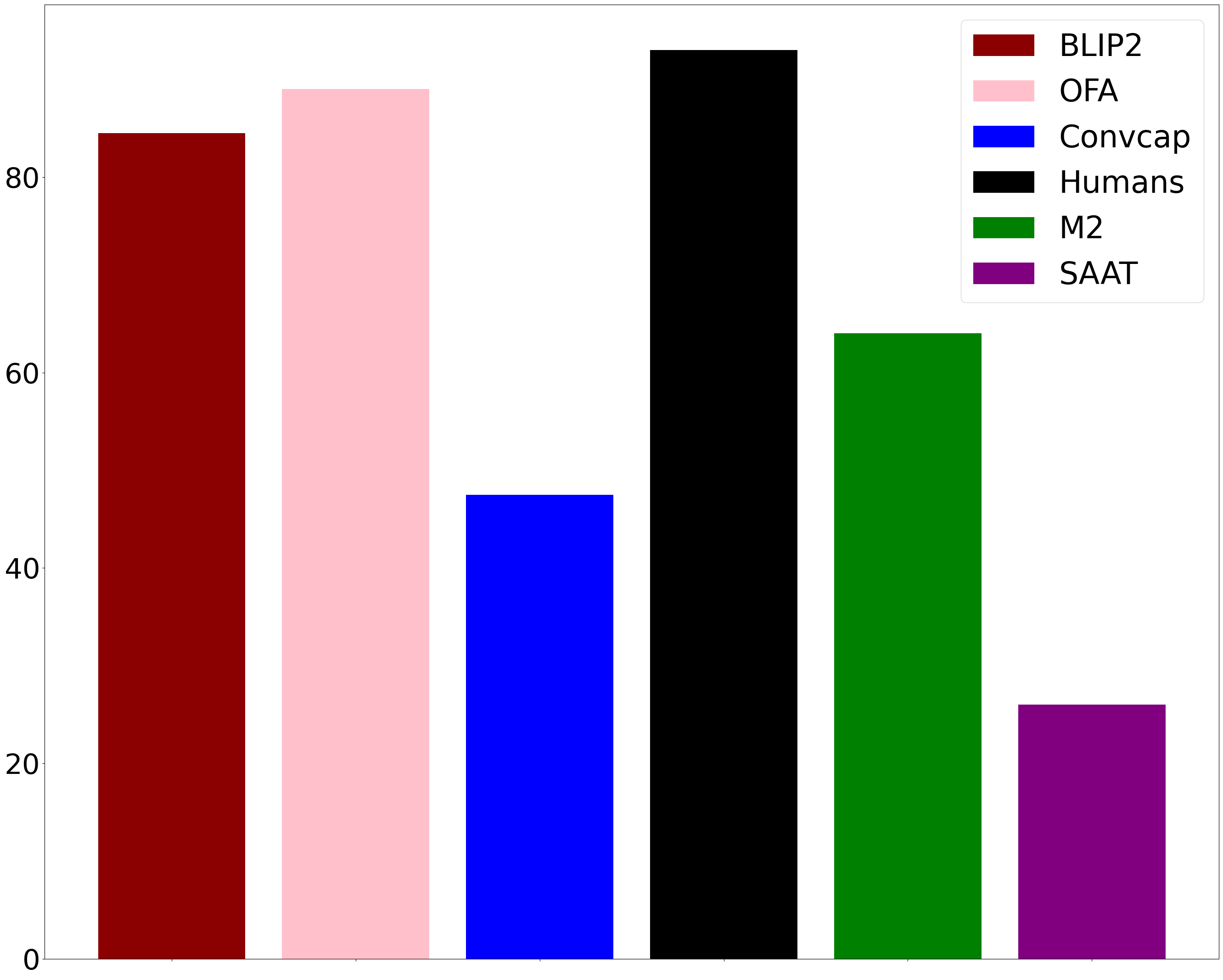}
    \caption{Voting scheme}
    \label{fig:voting}
\end{subfigure}
\hfill
\begin{subfigure}{0.45\textwidth}
    \includegraphics[width=\textwidth]{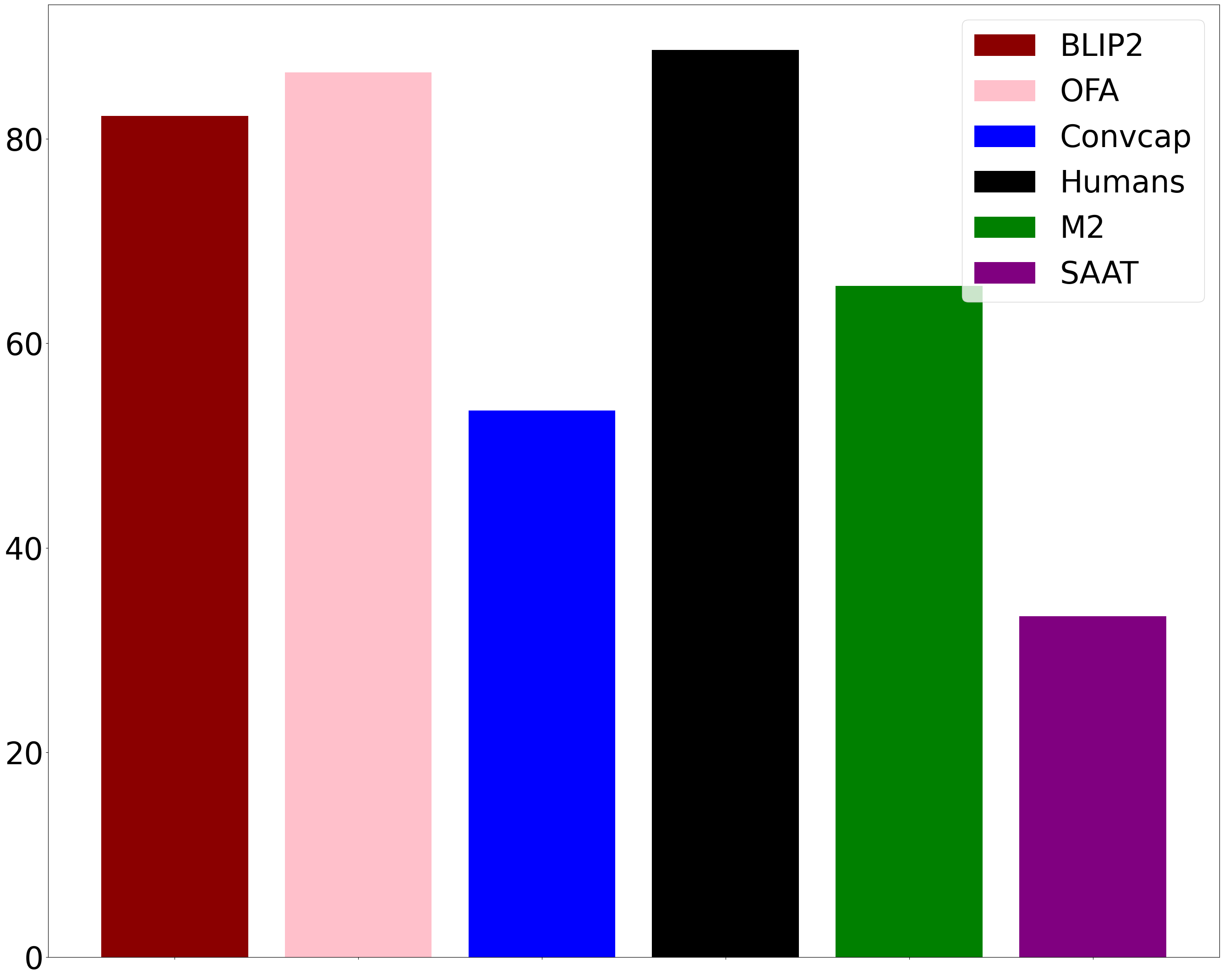}
    \caption{Mean scheme}
    \label{fig:mean}
\end{subfigure}
\hfill
\begin{subfigure}{0.45\textwidth}
    \includegraphics[width=\textwidth]{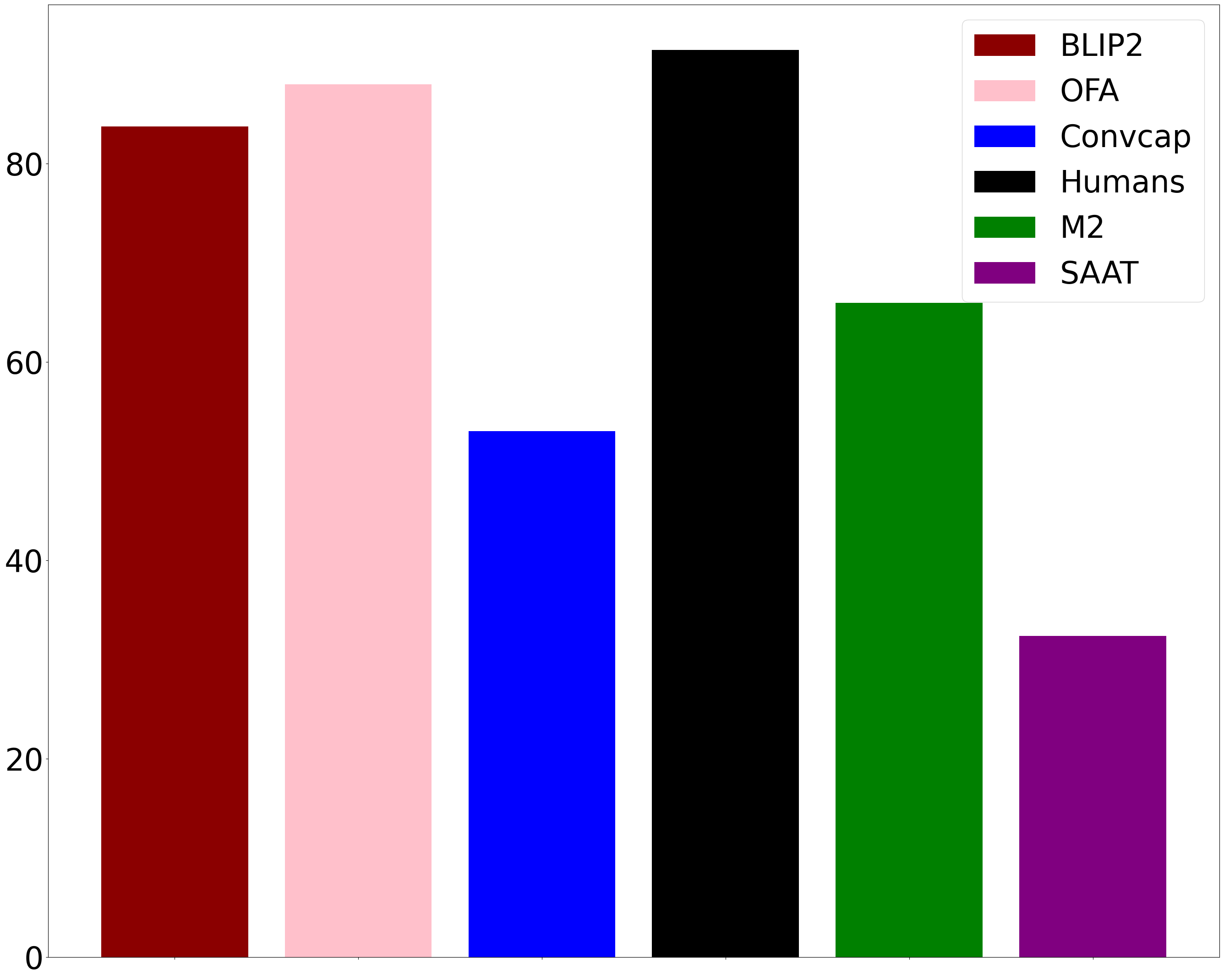}
    \caption{Eliminating max and min values, and sum up the rest}
    \label{fig:minmax}
\end{subfigure}
\caption{Overall score for each model, as well as the Reference of the MS-COCO dataset}
\label{fig:modelsevaluation}
\end{figure}

In a complementary way, Figure~\ref{fig:modelsevaluation} shows all the results for all the models as well as the reference from the MS-COCO dataset (named human-reference). It can be seen that the best image-caption from the models evaluated was the human-reference, followed by the OFA and BLIP2 models. The worst evaluated was the SAAT model, which is expected because it was one of the first models proposed in the literature, hence the oldest one in this comparison. This behavior is the same as voting and mean schemes, but the results are quite different in eliminating max and min values and summing the rest.

\begin{figure}[!h]
\centering
\begin{subfigure}{0.45\textwidth}
    \includegraphics[width=\textwidth]{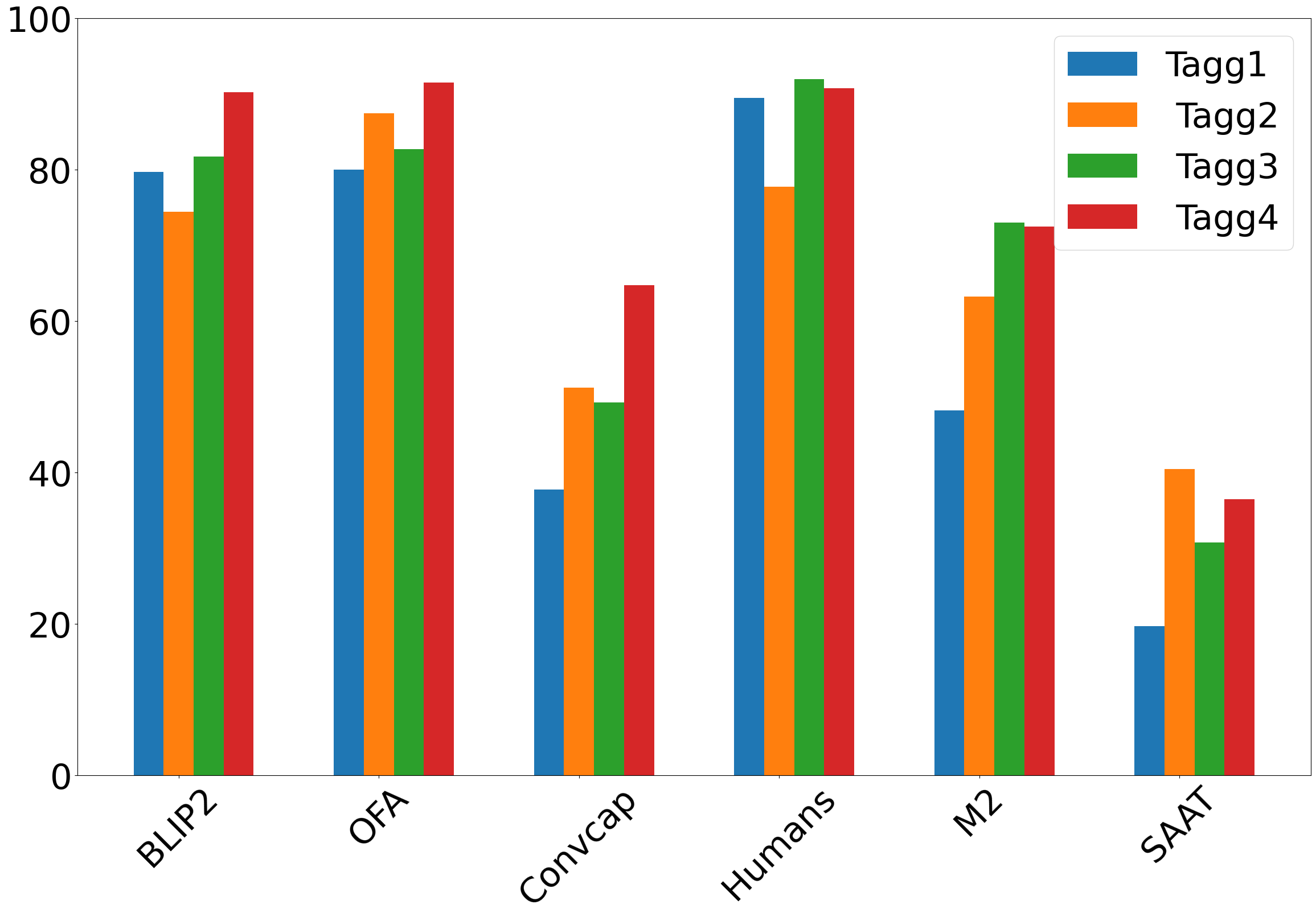}
    \caption{All Taggers (stage 1)}
    \label{fig:first}
\end{subfigure}
\hfill
\begin{subfigure}{0.45\textwidth}
    \includegraphics[width=\textwidth]{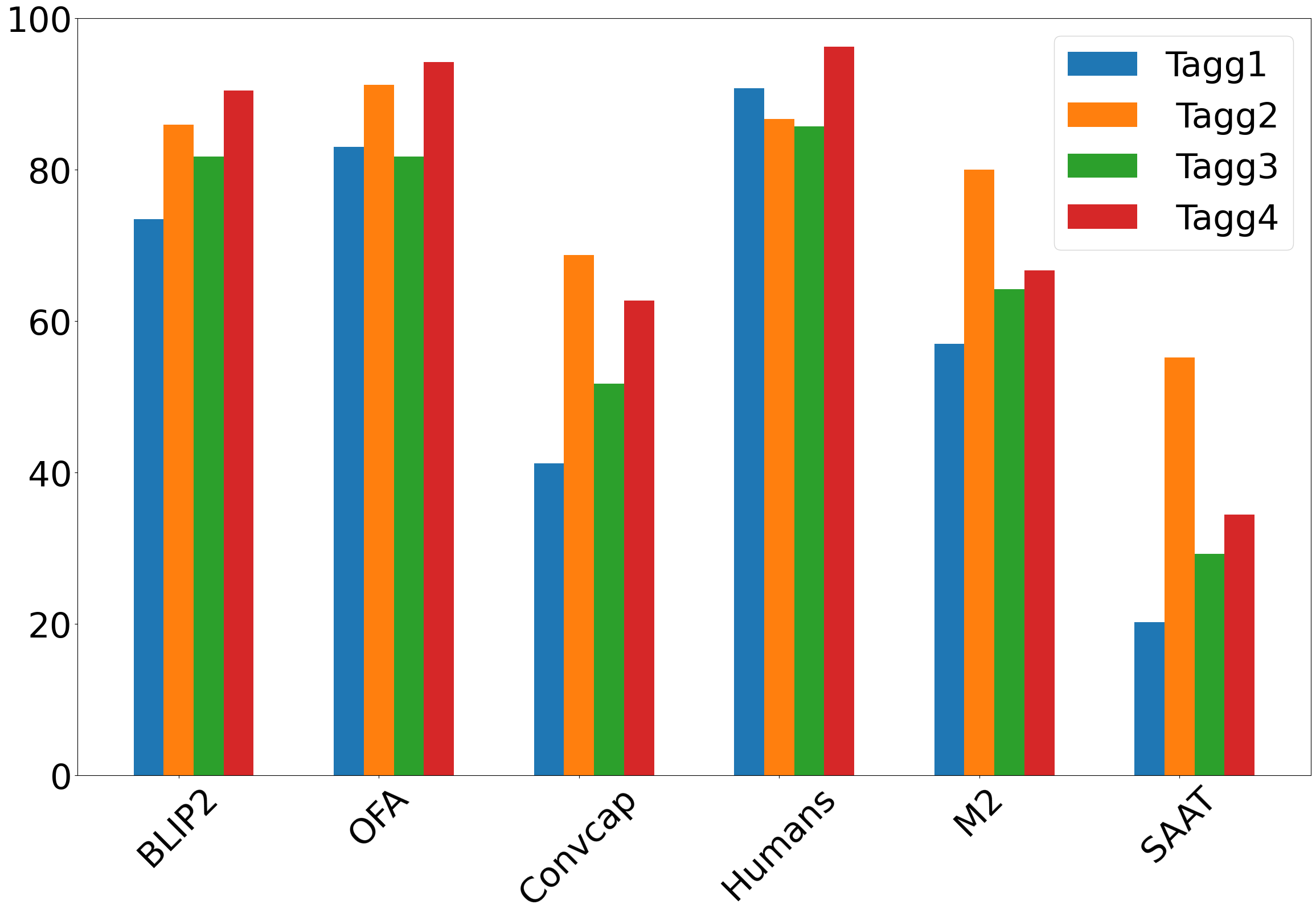}
    \caption{All Taggers (stage 2)}
    \label{fig:second}
\end{subfigure}
\caption{Scores assigned to each model's prediction by all the taggers in the two stages}
\label{fig:scoresfortagger}
\end{figure}

If we want to analyze how each Tagger evaluated each model, Figure~\ref{fig:scoresfortagger} shows the evaluation for all the models and each Tagger in the two evaluation stages. Figure~\ref{fig:first} shows the scores of each Tagger according to each model. Here, it can be seen that the highest values were achieved by human-generated captions in both stages.

From these figures, many aspects can be analyzed; for instance, in stage 1, the Tagger4 provides the highest value; in general, all the taggers maintained similar behavior in stage 1. However, in stage 2, Tagger2 changed his behavior to be more positive in this stage, which means higher scores compared to his performance in stage 1.

For the metric proposal and all the experiments, the mean score of all the human evaluations was considered as our dependent variable, or the $y$, for prediction.

\section{Proposal Metric}
\label{sec:proposal}

As a metric proposal\footnote{available in https://github.com/ResearchInfoGeo/VCRScore}, the idea was to generate a machine learning model replicating the task done by our four human taggers (on average), so this problem was tackled as a regression model where several inputs were used to approximate the score in a value range between 0 and 1, where 0 is the worst possible score and 1 the opposite.

We tried some ideas to build this metric proposal. All of them are explained in the following sections.

\subsection{Metrics used}
\label{sec:mostused}
\subsubsection{State-of-the-art metrics}
Some metrics are very well-known and widely used by the research community. We selected the standard metrics and tried some newer ones to compare them with our metric proposal.

As is known, the most used metrics to evaluate image captioning, are \emph{BLEU} \cite{Papineni02bleua}, \emph{METEOR}~\cite{Lavie2007}, \emph{ROUGE}~\cite{lin-och-2004-orange}, \emph{SPICE}~\cite{anderson2016spice}, and \emph{CIDEr}~\cite{vedantam2015cider}. Newer metrics have been also proposed, some of them are based on BERT~\cite{Devlin2019} and CLIP (Contrastive Language-Image Pre-Training)~\cite{radford2021learning}. From these new ones, we will use \emph{BertScore}~\cite{zhang2019bertscore}, \emph{CLIPScore}, and \emph{CLIPScore\_ref}~\cite{DBLP:journals/corr/abs-2103-00020}.

We will briefly introduce some important concepts from these metrics for space reasons, but if the reader wants to be deeper into them, please consult the corresponding reference.

\textbf{BLEU (Bilingual Evaluation Understudy)} is one of the most used metrics to evaluate image captioning models; it was originally designed for automatic translation evaluation; nevertheless, through the years, it has been adapted and intensely used in the evaluation of image captioning models.
BLEU is based on the co-occurrence of $n$-grams (such as words) analysis between the prediction and reference(s). Usually, this value of $n$ represents how a consecutive number of words will be matched to be considered a success, $n$ had been established with values from 1 to 4, and as it was expected when one used 4, lower results were achieved. In this comparison, we decide to use $4$-gram in BLEU.

\textbf{METEOR (Metric for Evaluation of Translation with Explicit ORdering)} tries to address some of the drawbacks of the BLEU metric. METEOR also looks for matching words considering precision and recall values for unigram matches and computes a harmonic mean with them. Also, it uses a penalty value to try to map the fewest feasible quantity of chunks or fragments in the sentence to unigrams; if the number of segments increases, the penalty increases as well.

\textbf{ROUGE (Recall-Oriented Understudy for Gisting Evaluation)}
measures the number of overlapping in term sequences ($n$-gram) and word pairs for the captions generated. This metric has four versions:  ROUGE-N, ROUGE-L, ROUGE-W, and ROUGE-S. We used ROUGE-L, which estimates the longest common subsequence.

\textbf{CIDEr (Consensus-based Image Description Evaluation)} tries to give a solution to the necessity for a metric that could better correlate with human judgment, employing a {\it consensus} between the several references in a corpus. CIDEr also uses the approach of $n$-grams, but in this case, firstly, all the words are converted into their root form. The TF-IDF (term frequency-inverse document frequency) is used to compute the final metric value.

\textbf{SPICE (Semantic Propositional Image Caption Evaluation)} does not use the $n$-gram approach; instead, it transforms both reference and prediction sentences into a scene graph. SPICE uses a pre-trained model based on the Stanford Scene Graph Parser~\cite{schuster2015generating}.
In the end, SPICE is the calculation of $F$-score reflecting the comparison between the two graphs (reference and prediction sentences).

As can be seen, the prior metrics are based on the $n$-gram matching approach, and the SPICE is different in that it relies on the lexical representation of the words. Nevertheless, some current metrics are based on more sophisticated approaches, looking more into a semantic space correlating the words by their meaning and not by how they are written.

\textbf{BertScore}, similar to previously described metrics, computes the similarity between the reference and the prediction sentences, but instead of doing exact matching, it computes token similarities using contextual embeddings (from the BERT model). With these embeddings, one can have a vector representation of two different words, and the cosine similarity is computed using the dot product of the unitary vectors. Then, BertScore is calculated through the maximum score of all the candidate tokens and calculating the precision, recall, and F1 scores.

\textbf{CLIPScore} uses the contrastive learning model called CLIP for its calculation. CLIP is a model trained with image-text pairs data using a dataset of 400M images with descriptions taken from the internet.
One of the things CLIPScore wants to avoid is human evaluation, which means CLIPScore wants to be a reference-free manner to evaluate the quality of a prediction. To evaluate the quality of a prediction, CLIPScore uses both the image and the predicted caption. Later, the cosine similarity of the resultant embeddings is calculated.
As an additional variety of CLIPScore, the authors in~\cite{DBLP:journals/corr/abs-2103-00020} proposed CLIPScore\_ref.
The latter is calculated using all the references to generate vector representations using CLIP's text transformer, and the score is computed as the harmonic mean of CLIPScore, as well as the maximal reference cosine similarity.

As can be seen, one can split the metrics described as those based on $n$grams matching or a more lexical approach and those based on embeddings (from images and texts) or a more semantic approach.
We consider that both approaches are necessary to calculate a score that correctly reflects the quality of a caption prediction.

\subsubsection{Model's embeddings as metrics}
Both BertScore and CLIPScore used pre-existing models trained on tasks that did not have image captioning in mind. The BERT model was trained to predict a masked token from its context. In the case of the CLIP, it was trained to match the embeddings of an image with the corresponding description. The embeddings produced by the BERT and CLIP models are distributed in an organized way. They have the property that close embeddings correspond to semantically related objects (text or images). This can be used to measure the similarity between different objects or representations, which can be applied to the problem of image captioning.

With that abstraction in mind, we tried to emulate the process using different models. The models we selected are described below.

\begin{itemize}
    \item \textbf{Multi-Caption-Image-Pairing} (MCIP)~\cite{DBLP:journals/corr/abs-2111-13122} is a fine-tuned CLIP where the image encoder is trained with a different loss function and a collection of related captions per image. The text encoder is kept frozen to maintain the alignment of the text and image embeddings. This fine-tuning is used to improve the model for the image retrieval task. In the image retrieval task, given a query (image of text), the model must find the $k$ most relevant images, where the similarity is measured with the cosine metric. This task is deeply related to image captioning, so we used it.
    \item \textbf{BertGrammar}. There are two important aspects that a good caption needs to have. First, the captions need to describe the most important objects of the image and their interactions. Another important aspect of the caption is its readability. The captions need to be well-written to others to interpret it. As our previous work shows~\cite{GONZALEZCHAVEZ2024117071}, only the BertScore takes grammar into account. In order to improve in this aspect, we used the BERT for Grammar Error Detection tool\footnote{\url{https://github.com/sunilchomal/GECwBERT/tree/master}} to add a score to the grammar of the captions. It is important to note that this metric does not need the image or the references to give a score, it only needs the predicted caption. This metric will be denoted as BertGrammar.
    \item \textbf{Vision-and-Language Transformer (ViLT)}, with the increasing popularity of the Vision-and-Language Transformer (ViLT) \cite{kim2021vilt}, we explore its representation to build a metric by training a model.
    The ViLT consists of a simple architecture for a vision-and-language model; the transformer extracts and processes visual features by making parameter efficiency, and the computational size of textual and visual embed is lower than others and employs the process in modeling modality interactions~\cite{kim2021vilt}.
    For this work, the ViLT obtains a distance score between the model encoded by the image and the caption. After the regression model utilizes this result as a feature to obtain a final score for the pair image-caption, other features are also used.
\end{itemize}

\subsection{Precision and Recall calculation}
As a part of our proposal inspired by~\cite{kasai-etal-2022-transparent} we decided to use two well-known concepts, \emph{Precision} and \emph{Recall} metrics. As one knows, these metrics are widely used in any machine learning problem.
Kasai et al.~\cite{kasai-etal-2022-transparent} proposed the THUMB, a rubric-based human evaluation protocol for image captioning models. They proposed a protocol for evaluating models done by humans and also analyzed how automatic metrics correlate with these human scores.
Hence, inspired by their ideas, we calculate our versions of \emph{Precision} and \emph{Recall}.
In this case, \emph{Precision} measures the accuracy of the information in a caption, and \emph{Recall} assesses how much of the salient information in the image is covered.
Set our set of reference words as $S$ with dimension $n$ (that means number of words), and $Z$ and $i$ our set of caption prediction words and the corresponding dimension, respectively.
For our calculation, we first generate an amplified set called $pool$, which is the concatenation of all the references $S$ plus additional words generated with two object detection models, Yolov3~\cite{han2021you} and DEtection TRansformer or DETR \cite{carion2020end}. These two models provided us with a set of the names of the detected objects in the image. We concatenated all of them, keeping the unique ones, and in this way, the $pool$ was built.
$pool$ is the augmented set of words related to the image content, considering both the captions references and objects found in the image.

Considering the prior, we define $r$, \emph{precision} and \emph{recall} as follows:

\begin{equation}
    r = \left|Z \cap pool \right|,
\end{equation}
\begin{equation}
    precision = \frac{r*100}{\left| Z \right|},
\end{equation}

\begin{equation}
    recall = \frac{r*100}{\left| pool \right|},
\end{equation}

where $r$ is the number of words in both prediction and $pool$, then, \emph{precision} gives us the proportion of words of the prediction being in the $pool$ according to the size of our prediction sentence $i$.
\emph{Recall} gives the proportion of words of the prediction being in the $pool$ according to the size of the $pool$.
In this sense, \emph{recall} tends to be lower than \emph{precision}. Both values are between 0 and 1.
We use these two values to complement the input necessary to train our metric; results are shown in the experiments section (see Section \ref{sec:expandana}).

\subsection{Gradient Boosting Regressor}
The regression algorithm Gradient Boosting Regressor (GBR) was used. Some training processes, using a combination of different variables obtained from the Vision Transformer, CLIP, and the recall, previously explained, were carried out.

Friedman \cite{friedman2001greedy} proposes extending the boosting algorithm to competitive and robust regression and classification problems. This technique allows the optimization of the loss function. It employs gradient descent to minimize and build a prediction model based on common decision trees.

The GBR is trained with features related to precision, recall, and the estimated score of the transformer (ViLT) and CLIP. The target is the human evaluation in the pair image-caption dataset, where the machine-learning regressor is applied.

\subsection{Spearman Correlation}
The Spearman correlation coefficient was employed to evaluate how close the automatic metrics to the human score are in rankings positions as well as in the produced scores.
Usually, the Spearman rank correlation coefficient~\cite{spearman} is used as a statistical test to determine if there exists a monotonic relationship between two variables. That means that the Spearman coefficient evaluates the strength and direction of the relationship between two variables. It takes values between -1 and 1, where 1 means an exact proportional correlation and -1 is an inverse correlation.

The equation that calculates the Spearman correlation $\rho_s$ is:

\begin{equation}
\rho_{s}=1-{\frac {6\sum d_{i}^{2}}{n(n^{2}-1)}\,\,},
\end{equation}

\noindent where $d_i$ is the difference between the two ranks of each observation and $n$ is the number of observations.

\section{Experiments and analysis}
\label{sec:expandana}

This section will detail all the datasets used since we used both types, the built by the taggers (procedure explained in Section~\ref{sec:tagging}) and the ones acquired by other authors.
Also, the results obtained with our metric proposal and those achieved by the current and previously explained metrics were compared.

\subsection{Data}
\label{sec:data}
To train our metric, we use two kinds of datasets, first our generated dataset, which results from the human-tagging process (explain in Section~\ref{sec:tagging}). Also, we collected other valuable datasets such as VICR~\cite{NEURIPS2023_c0b91f9a} where the scores were taken in a gamified fashion; that is, participants were encouraged to guess the consensus on each particular caption. The dataset contains $9,990$ unique images with $15,646$ pairs of image-caption evaluations. The evaluation took values from 1 to 5, Table~\ref{tab:VICR} described the meaning of each value.
\begin{table}[!h]
    \centering
    \caption{VICR rating scale descriptions}
    \vspace{-.3cm}
    \label{tab:VICR}
    \resizebox{11cm}{!}{
    \begin{tabular}{|c|p{4in}|}
        \hline
        Rating & Meaning \\ \hline
        1 & Objects are incorrectly identified. The caption gives the wrong idea about what is happening
in the image.\\
        2 & Objects are partially correctly identified with some errors, but the caption is accurate enough
to give an idea of what is happening in the image. The caption identifies most of the objects
but might not identify everything. There is no interpretation of what anything means.\\
        3 & Relevant objects are correctly identified. The caption describes what is seen but not where
objects are in space. There is no description of the overall setting and no interpretation of an
event.\\
        4 & Objects and/or a general scene and/or an action are correctly identified but not every element
is completely identified. The caption describes what is seen and where things are in space.
There is no interpretation of an event. \\
        5 & Objects, a general scene, and actions are correctly identified if present in the image. The
caption describes what is seen and where things are in space. Interpretation of the overall setting
and/or event is included. \\ \hline
    \end{tabular}
    }

\end{table}

The Flickr8k-Expert~\cite{10.5555/2566972.2566993} was also used, which contains $1,000$ images and $5,822$ evaluations. The scores range from 1 to 4; please refer to Table~\ref{tab:Flickr8k-Exp} to see the meaning. A similar dataset is Flickr8k-CF, which is presented in the same article. It includes $1,000$ images and $47,829$ pair evaluations. This time, the task was binary to indicate whether the caption described the image (Yes) or not (No). The final score was the percentage of Yes's obtained from 3 reviewers.

\begin{table}[!h]
    \centering
    \caption{Flickr8k-Expert rating scale descriptions}
    \vspace{-.3cm}
    \label{tab:Flickr8k-Exp}
     \resizebox{.6\textwidth}{!}{
    \begin{tabular}{|c|l|}
        \hline
        Rating & Meaning \\ \hline
        1 & Is unrelated to the image.\\
        2 & Is somewhat related to the image.\\
        3 & Describe the image with minor errors.\\
        4 & Describe the image without any errors. \\ \hline
    \end{tabular}
    }

\end{table}

The last dataset used was the Composite~\cite{ADITYA201833}, which includes $997$ images from Flickr8k and $2,973$ pair evaluations, from Flickr30k another $991$ images and $3,964$ evaluations, and MS-COCO (VAL 2014) $2,007$ unique images and $8,028$ ratings. The ranking scale description is in Table~\ref{tab:Comp}.

\begin{table}[!h]
    \centering
    \caption{Composite rating scale for Relevance and Thoroughness.}
    \vspace{-.3cm}
    \label{tab:Comp}
         \resizebox{.7\textwidth}{!}{
    \begin{tabular}{|c|l|}
        \hline
        Relevance & Meaning \\ \hline
        1 & The description has
no relevance.\\
        2 & The description has only weak relevance.\\
        3 & The description has some relevance\\
        4 & The description relates closely \\
        5 & The description relates perfectly\\ \hline
        Thoroughness & Meaning \\ \hline
        1 & The description covers nothing.\\
        2 & The description covers minor aspects.\\
        3 & The description covers some aspects.\\
        4 & The description covers many aspects. \\
        5 & The description covers almost every aspect.\\ \hline
    \end{tabular}
    }

\end{table}

\begin{figure}[!h]
    \centering
    \begin{subfigure}{.45\textwidth}
        \centering
        \includegraphics[width=0.4\linewidth]{ex-vicr.jpg}
        \caption{A women who is standing by a barb wire fence. (0.65)}
    \end{subfigure}
    \begin{subfigure}{.45\textwidth}
        \centering
        \includegraphics[width=0.4\linewidth]{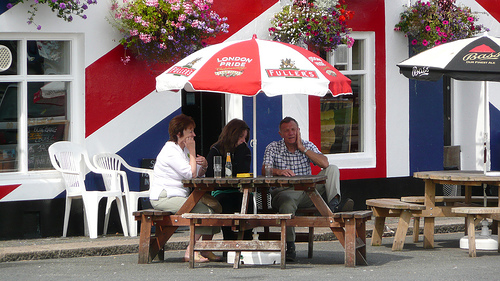}
        \caption{a young man jumping a back flip off of a concrete wall. (0.11)}
    \end{subfigure}
    \begin{subfigure}{.45\textwidth}
        \centering
        \includegraphics[width=0.4\linewidth]{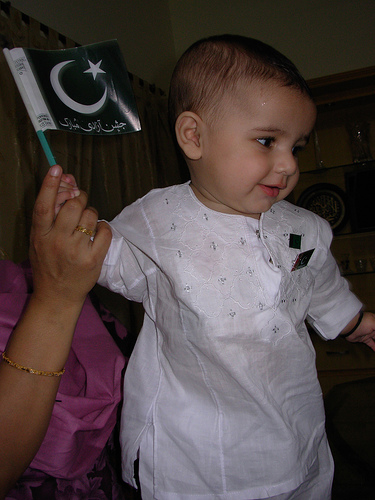}
        \caption{A baby wearing a white gown waves a Muslim flag. (1.0)}
    \end{subfigure}
    \begin{subfigure}{.45\textwidth}
        \centering
        \includegraphics[width=0.4\linewidth]{ex-comp.jpg}
        \caption{person is smiling in the scene. (0.4)}
    \end{subfigure}

    \caption{Examples of captions in the different datasets. (a) is for VICR, (b) is Flickr8k-Expert, (c) is Flickr8k-CF, and (d) Composite. The normalized score is in parentheses.}
    \label{fig:caption-examples}
\end{figure}

The scores from all the datasets were normalized to have values from 0 to 1. Figure~\ref{fig:caption-examples} shows some examples of captions and scores for the datasets VICR, Flickr8k-Expert, Flickr8k-CF, and Composite.

Table~\ref{tab:overalldata} shows the number of samples on each dataset. First is our generated dataset, composed of $600$ samples. Secondly, as explained before, is the set of public datasets of the state-of-the-art, which in total provides a higher number of samples, but we decided to eliminate those samples with 0's values on their score since they were considered not helpful to our model generation. So, we reduced the second dataset from $81,289$ to $59,315$, meaning a total of $21,974$ samples with an evaluation score of 0 were removed. Also, it is important to say that in the case of our generated dataset, it was split into 30\% of train and 70\% of test because of its size, and the opposite way was done with the external datasets, which was divided into 70\% of training and 30\% of test.

\begin{table}[!b]
    \centering
    \begin{tabular}{|c|c|c|c|}
    \hline
    \bf{Dataset} &  \bf{Total} &  \bf{Train set} & \bf{Test set} \\ \hline
      Our Generated dataset   & 600 & 240 (30\%) & 360 (70\%) \\ \hline
        VICR  & 15,646  & - & - \\ \hline
        Flickr8k-Expert & 5,822 & - & - \\ \hline
        Flickr8k-CF &47,829& - & - \\ \hline
        Composite  &11,992&- &  - \\  \hline
        Total &59,315 & 41,521  (70\%) & 17,794(30\%) \\ \hline
    \end{tabular}
    \caption{Total of data used for experiments. Our build dataset was split into 30-70 percent of train and test, respectively. In the case of the other datasets, as a whole, it was split into 70-30 percent train and test, respectively.}
    \label{tab:overalldata}
\end{table}

\subsection{Computed scores of each metric}
As mentioned, our analysis was conducted using a set of image captioning models and several performance metrics.
The metrics were explained in Section~\ref{sec:mostused}. The model evaluated were OFA~\cite{wang2022ofa}, BLIP2~\cite{li2023blip2, blip2}, Meshed-Memory Transformer (M$^2$)~\cite{cornia2020m2}, Convcap~\cite{AnejaConvImgCap17},  and Show, Attend and Tell~\cite{pmlr-v37-xuc15} (SAAT). Additionally, we used the first reference of the MS-COCO dataset, which was considered a human-generated caption (Humans).
Table~\ref{tab:resultadosmetricas_1} and \ref{tab:resultadosmetricas_2} show the scores produced by each metric computed for each model tested. The results were split; Table~\ref {tab:resultadosmetricas_1} shows those metrics based on lexical approaches as well as they are the old ones, which are BLEU, ROUGE, SPICE, CIDEr, and METEOR. Also, the human score obtained by the tagging procedure is shown in column Tagging. According to the scores, the best model is OFA (in bold), except for the human score (tagging).

Table~\ref{tab:resultadosmetricas_2} shows the metrics based on embedding or trained models (included the proposed in this work), and also they are relatively the newer ones. Both tables show the results from the two kinds of datasets used and described in Section~\ref{sec:data}. According to these scores, the best model is again OFA.

\begin{table}[!h]
\centering
    \caption{Metrics results (lexical-based metrics). EData means External Dataset. Tagging means our human score}
    \vspace{-0.3cm}
    \label{tab:resultadosmetricas_1}
 \resizebox{0.8\textwidth}{!}{
\begin{tabular}{lrrrrrr}
\toprule
Model & BLEU & ROUGE & METEOR & CIDEr& SPICE & Tagging  \\
\midrule
Humans & 0.087 & 0.458 & 0.239 &1.521 & 0.258  & \bf{0.886} \\
\bf{OFA} & \bf{0.290} & \bf{0.597} & 0.325 &\bf{1.918} &\bf{0.276} & 0.865\\
BLIP2 & 0.212 & 0.548 & 0.283 &1.751 &0.267  & 0.822\\
M$^2$ & 0.245 & 0.564 & \bf{0.521} & 1.800 & 0.258 & 0.656\\
Convcap & 0.205 & 0.550 & 0.275 & 1.396 & 0.187 &0.534\\
SAAT & 0.013 & 0.312 & 0.152& 0.740 & 0.136 &  0.333\\
\hline
EData &  0.068  & 0.373  & 0.352 & 1.522 &  0.231 & 0.721\\ \hline

\bottomrule
\end{tabular}
}
\end{table}

\begin{table}[!h]
\caption{Metrics results (embeddings-based metrics). MCIPScore and MCIPScore\_ref stand for retrieval options of CLIP. EData means External Dataset. VCRScore is our proposal metric. BertG is BertGrammar.}
    \vspace{-0.3cm}
    \label{tab:resultadosmetricas_2}
 \resizebox{\textwidth}{!}{
\begin{tabular}{lrrrrrrrrr}
\toprule
Model & BertScore & CLIPScore & CLIPScore\_ref & VCRScore & MCIPScore & MCIPScore\_ref & ViLT & BertG\\
\midrule
Humans & 0.93 & 0.76 & 0.80 & 0.91 & 0.52 & 0.65 & 0.81&  0.76 \\
\bf{OFA} & \bf{0.94} & \bf{0.78} & \bf{0.83} & \bf{0.92} &\bf{0.53} & \bf{0.66} & \bf{0.83} &0.91 \\
BLIP2 & 0.94 & 0.76 & 0.82 & 0.91 & 0.52 & 0.65 & 0.80& 0.91 \\
M$^2$ & 0.94 & 0.39 & 0.53 & 0.88 & 0.49 & 0.63 & 0.80 & 0.86 \\
Convcap & 0.93 & 0.72 & 0.77 & 0.78 & 0.44 & 0.57 & 0.69& \bf{0.93} \\
SAAT & 0.89 & 0.68 & 0.72 & 0.67 & 0.39 & 0.51 & 0.57& 0.57 \\
\hline \hline
EData & 0.91 & 0.69 & 0.73 & 0.72  & 0.53 & 0.42 & 0.62 &0.76\\
\bottomrule
\end{tabular}

}
\end{table}

These scores reflect the range of values of each tested metric as well as their variation across good and less good results of the image captioning models tested.

\subsection{Spearman correlation values with Human Evaluation}
\label{sec:correlationmetrisandmodels}

Here, we want to know how different the scores of the automatic metrics are from those given by humans. It is important to remember that in our dataset, we used 100 distinct images and the generated captions of the six different models (including humans). Then, we have 600 image-caption pairs in total, of which 360 were used as a test. From the external data, we have 17,794 additional image-caption pairs for testing. In this test set, we compared the correlation of the metrics' scores with our human tagging. We used the Spearman correlation to find the metric most similar to how humans judge.

The results are shown in Table~\ref{tab:metrics-correlation}. There, one can see that lexical-based metrics get small correlations with humans. The correlation increases using the BertScore, CLIPScore, and CLIPScore\_ref. The correlation gets even higher for the MCIP version of CLIP.
Something interesting happened with BertGrammar, the correlation with all the models in our dataset is high but with humans, where some of the captions get values close to zero, this can be seen in Figure~\ref{bertvsbergrammar}.
This suggests that, humans are more interested in the actual description than the grammar. We can also see that the ViLT metric correlates strongly with humans. Our metric proposal, the VCRScore, is highly correlated with human tagging.

\begin{table}[!h]
    \centering
    \begin{tabular}{|c|c|c|}\hline
        \bf{Metric} &\bf{Our dataset}  & \bf{EData}\\ \hline
       BLEU  & 0.174 & 0.218 \\ \hline
       ROUGE  & 0.391 & 0.462 \\ \hline
       METEOR & 0.376 & 0.468 \\ \hline
       SPICE  & 0.350 &	0.449   \\ \hline
       CIDEr &  0.360 &	0.522 \\ \hline
       BertScore &0.472 & 0.431 \\ \hline
       CLIPScore &0.506 & 0.652\\ \hline
       CLIPScore\_ref & 0.578 & 0.693  \\ \hline
       MCIPScore&  0.586& 0.705 \\ \hline
       MCIPScore\_ref&0.620 & 0.716 \\ \hline
       BertGrammar & 0.468& 0.711\\ \hline
       ViLT &0.508  & 0.709\\ \hline
       \textbf{VCRScore}& \bf{0.649} &  \bf{0.743} \\ \hline
    \end{tabular}
    \caption{Spearman correlation values between the metrics and the human evaluation on both our generated and external datasets (EData) for the test partition.}
    \label{tab:metrics-correlation}
\end{table}

\subsection{Evaluation of the Image Captioning Models}

\begin{figure}[!h]
    \centering
    \includegraphics[width=1\linewidth]{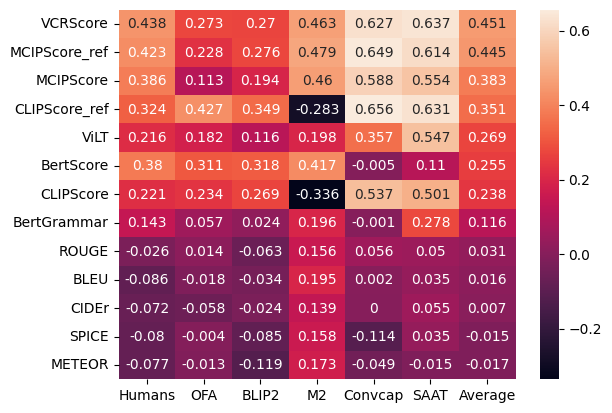}
    \caption{HeatMap of the correlation results of all the metrics evaluated}
    \label{fig:heatcorr}
\end{figure}

Figure~\ref{fig:heatcorr} shows the Spearman correlation between each model evaluated by humans and all metrics considered, expressed as a heatmap, where higher values indicate higher correlation and lower the opposite case where the values go from darker to lighter colors passing through red values which are medium (from the range of -1 to 1). Also, the average value is computed for comparison purposes.

Although, on average, the correlations are low with some models, and also low in average, one can see the worst correlated metrics were BLEU, ROUGE, and, even with low negative scores, METEOR and SPICE.
On average, the most correlated metrics are the BertScore, the two variations of CLIPScore, as well as the variations of CLIP for information retrieval (MCIPScore and MCIPScore\_ref).
However, the most correlated metric is the proposed VCRScore, with a correlation of 0.45 on average.

\begin{figure}[!h]
\centering
\begin{subfigure}{.45\textwidth}
    \centering
    \includegraphics[width=1.1\linewidth]{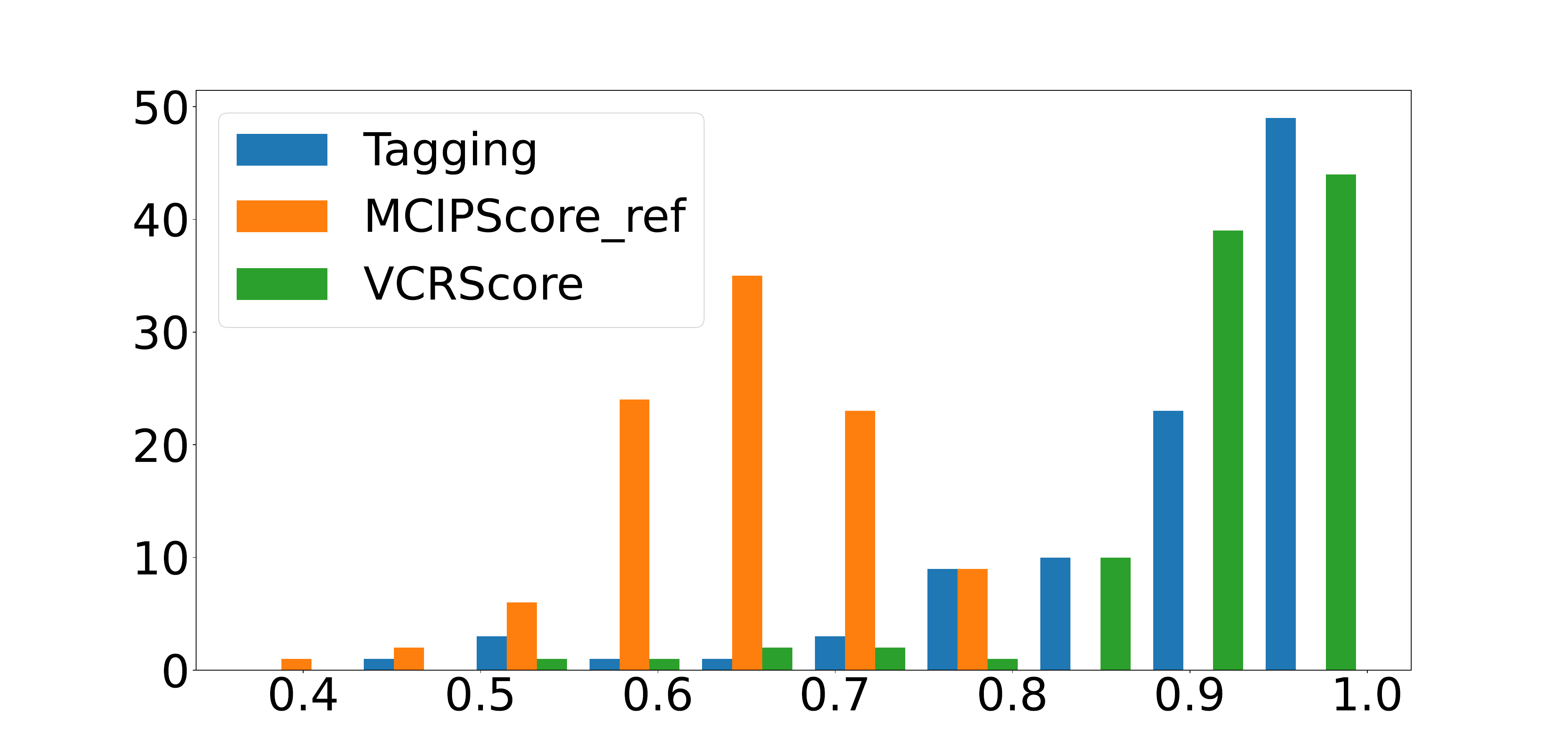}
    \caption{Humans}
\end{subfigure}
\begin{subfigure}{.45\textwidth}
    \centering
    \includegraphics[width=1.1\linewidth]{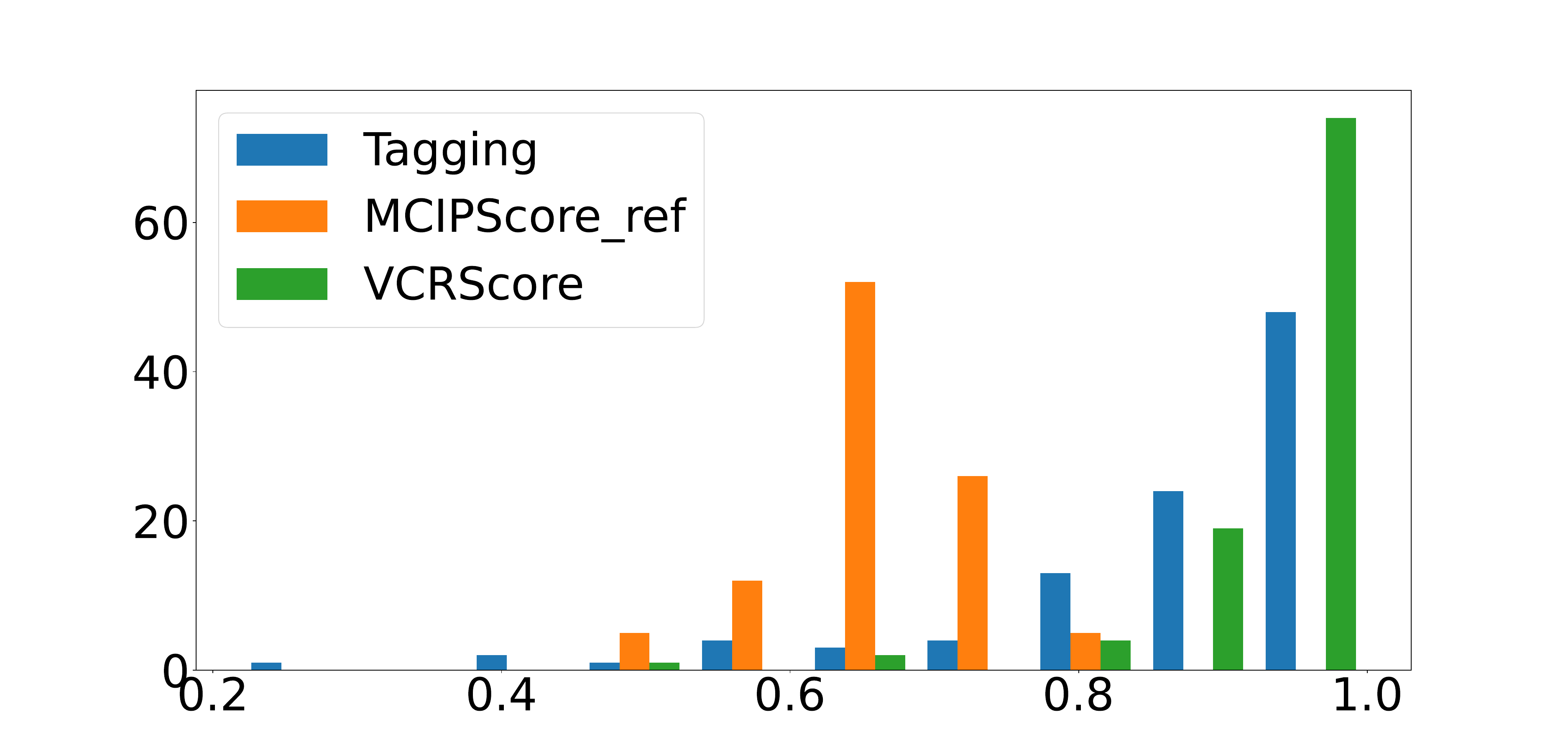}
    \caption{OFA}
\end{subfigure}
\\
\begin{subfigure}{.45\textwidth}
    \centering
    \includegraphics[width=1.1\linewidth]{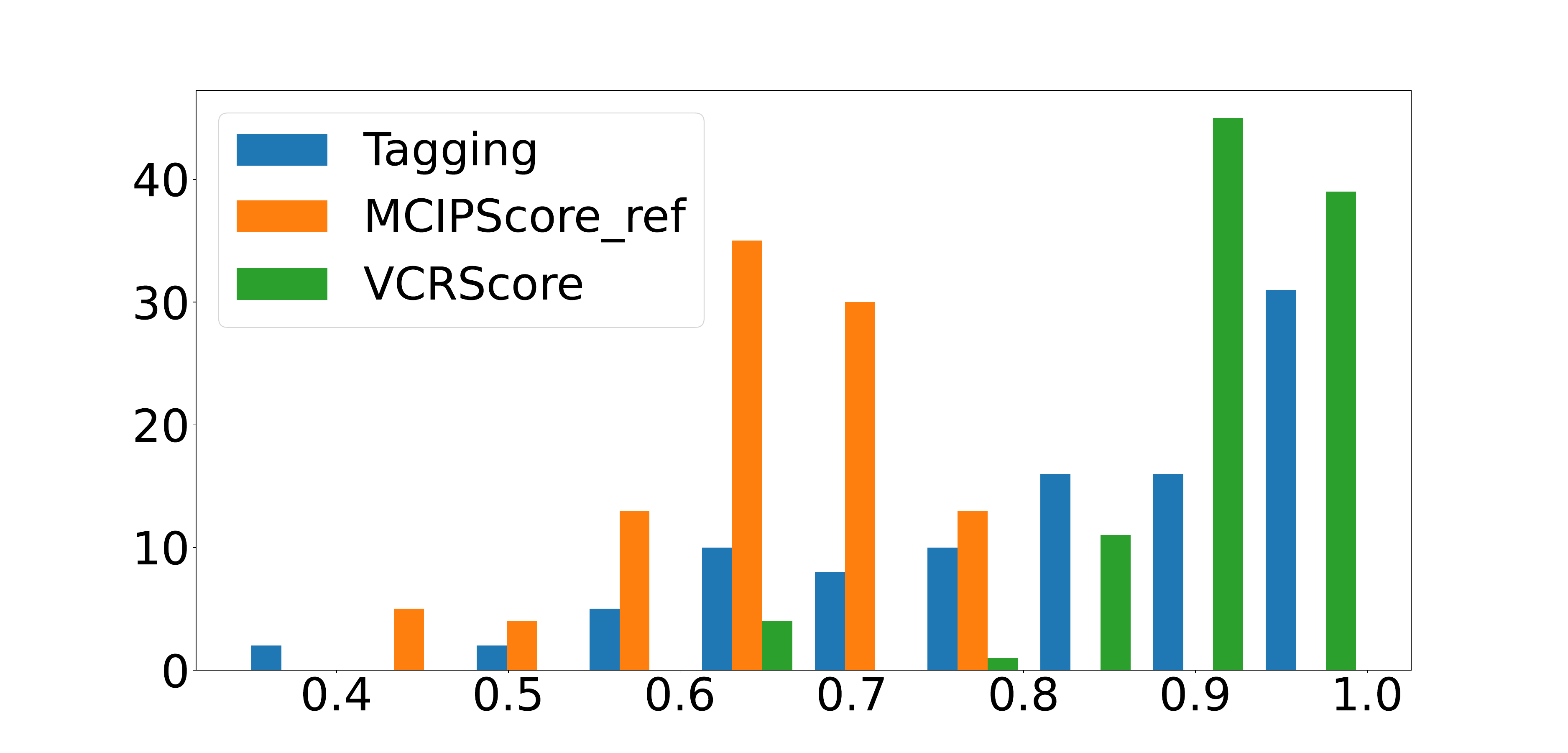}
    \caption{BLIP2}
\end{subfigure}
\begin{subfigure}{.45\textwidth}
    \centering
    \includegraphics[width=1.1\linewidth]{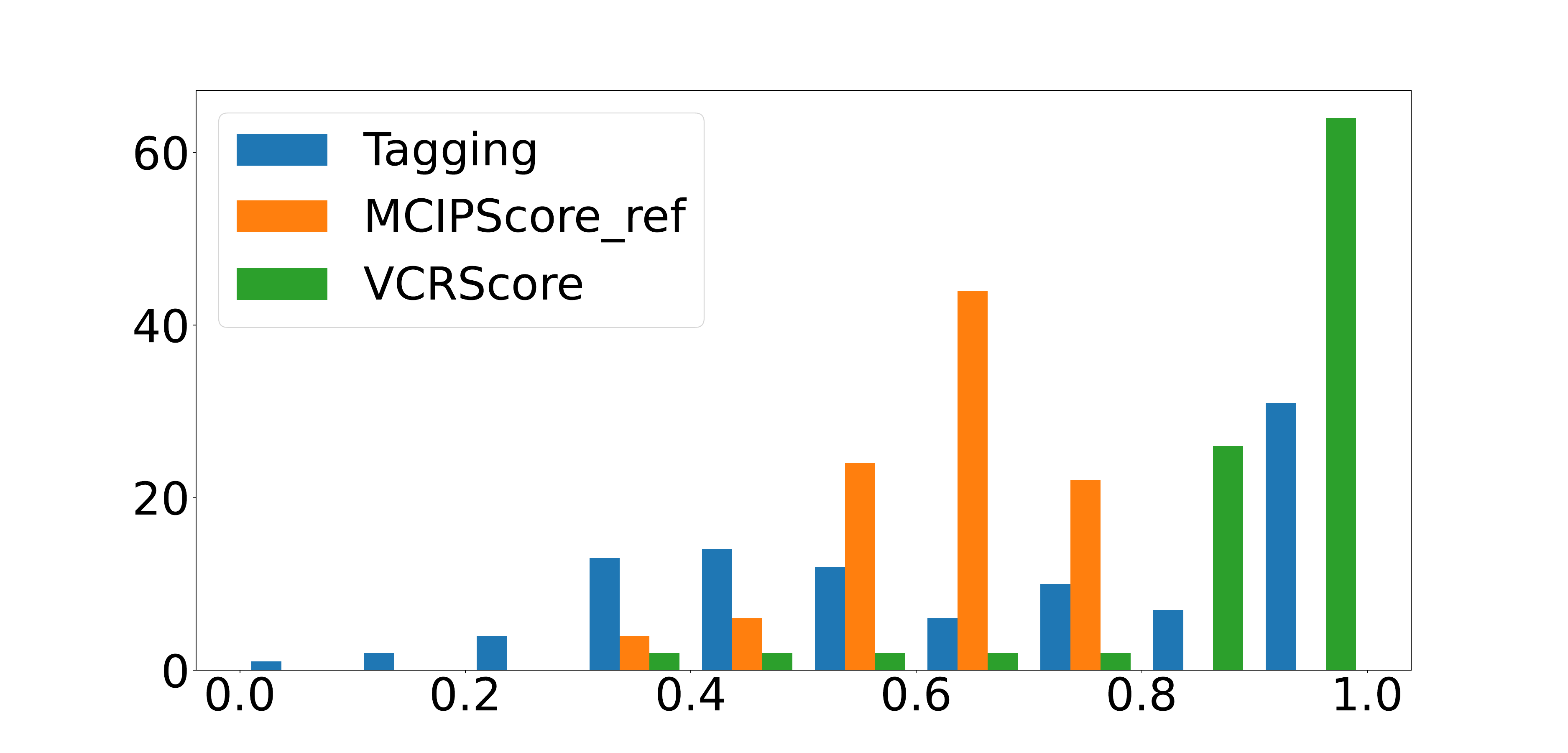}
    \caption{M$^2$}
\end{subfigure}
\\
\begin{subfigure}{.45\textwidth}
    \centering
    \includegraphics[width=1.1\linewidth]{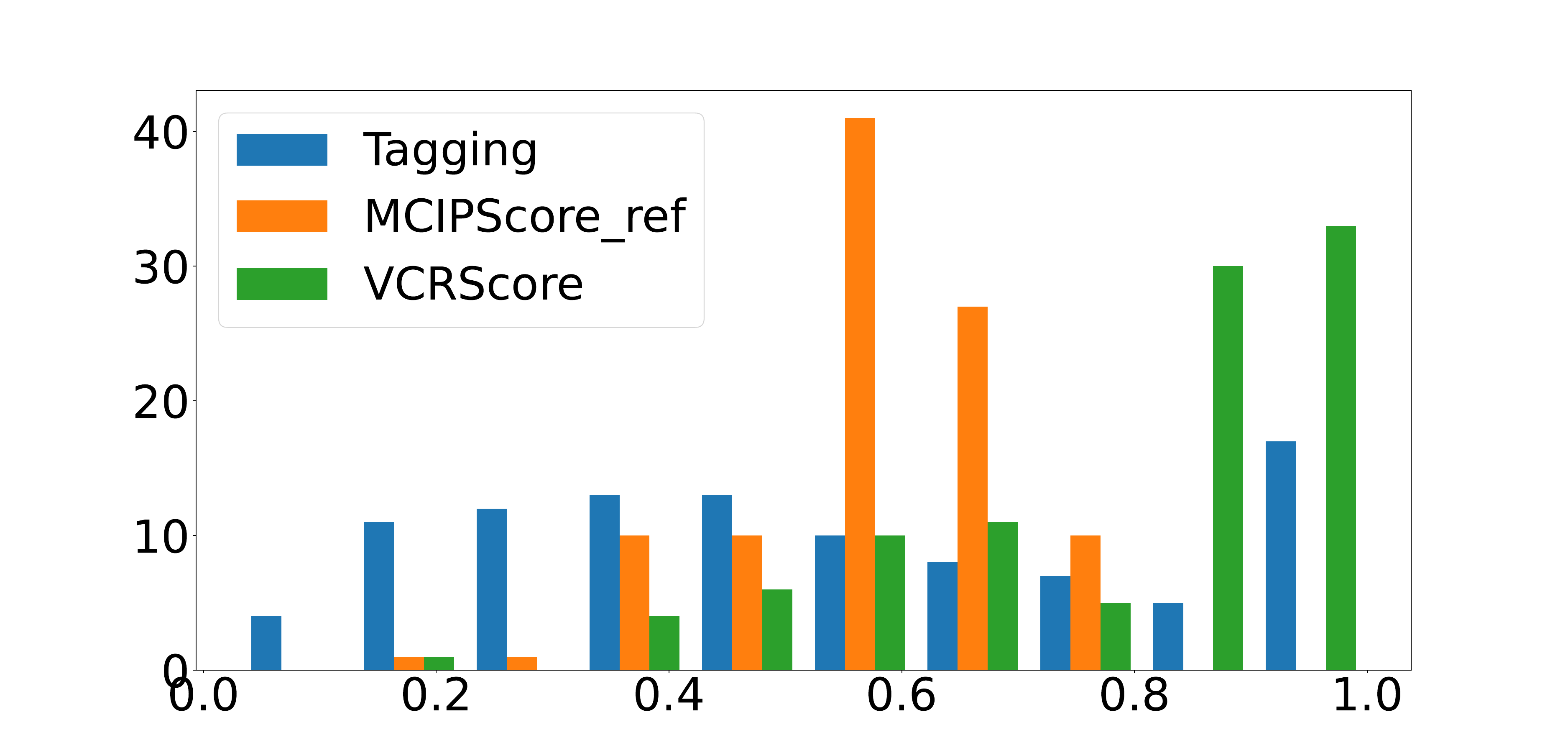}
    \caption{Convcap}
\end{subfigure}
\begin{subfigure}{.45\textwidth}
    \centering
    \includegraphics[width=1.1\linewidth]{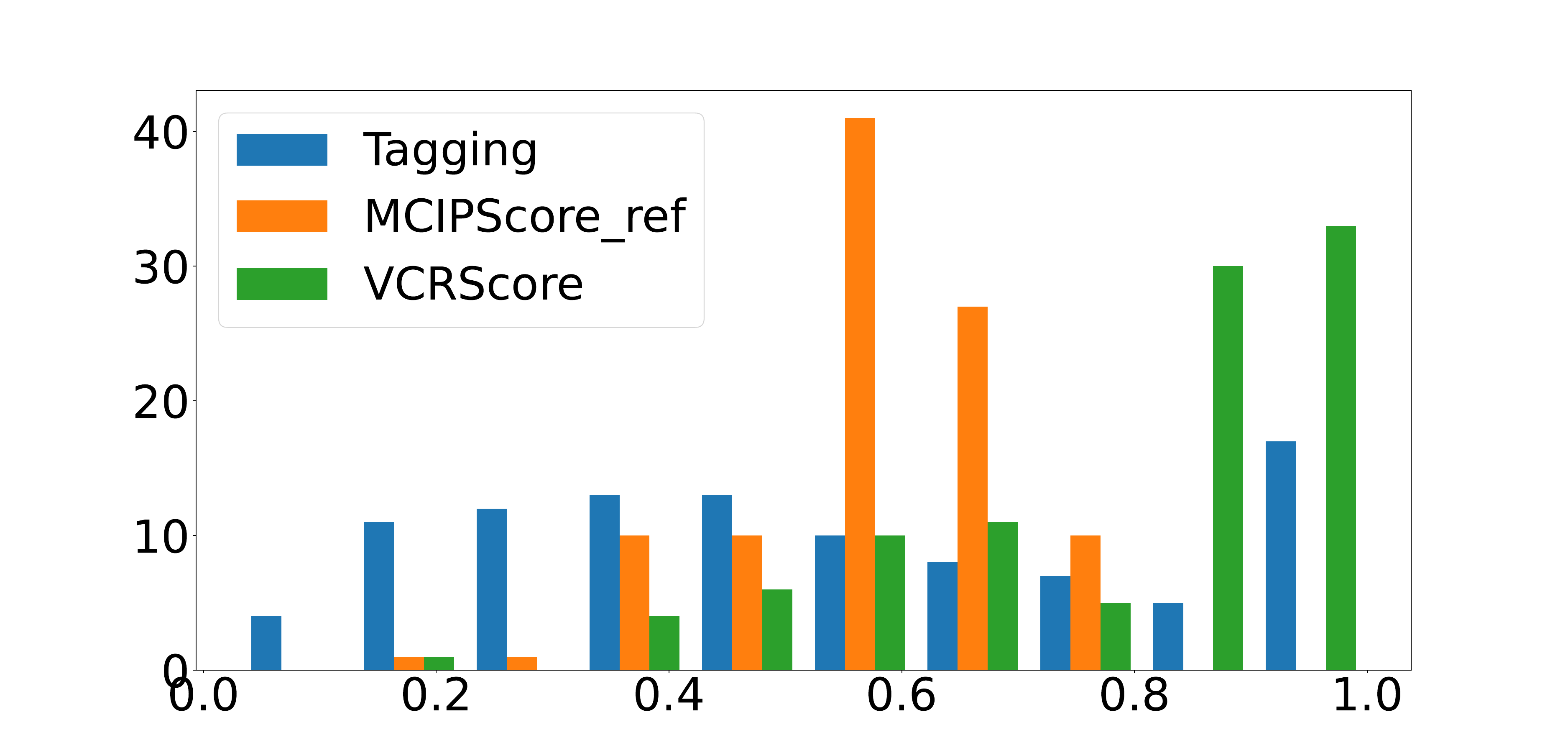}
    \caption{Show, Attend and Tell (SAAT)}
\end{subfigure}
\caption{Scores of the models with proposed metric and the human score (Tagging) and the MCIPScore\_ref }
\label{proposedvshumansscore}
\end{figure}

In the next series of figures, we want to show how similar the distributions of the automatic metrics and our tagging process are. These figures show two metrics at a time against human evaluation for comparison purposes. The plots were divided by model to facilitate interpretability.
Figure~\ref{proposedvshumansscore} plots the distributions of our tagging process, our proposed metric, and the second most correlated metric: MCIPScore\_ref. One can see that our proposal has a more similar histogram with human tagging than the second most correlated metric; this is for Humans, OFA, BLIP2, and Convcap models. For M$^2$, one can see the humans give more scores in the rank $[0.2, 0.6]$ and a few high scores, which contrasts with our VCRScore, which gives mostly high scores. Both automatic metrics fail to follow the human distribution for the SAAT model. In this particular case, one can see how the human tagging produced very low scores, which was not captured by any of the two metrics compared in this figure.

\begin{figure}[!h]
\centering
\begin{subfigure}{.45\textwidth}
    \centering
    \includegraphics[width=1.1\linewidth]{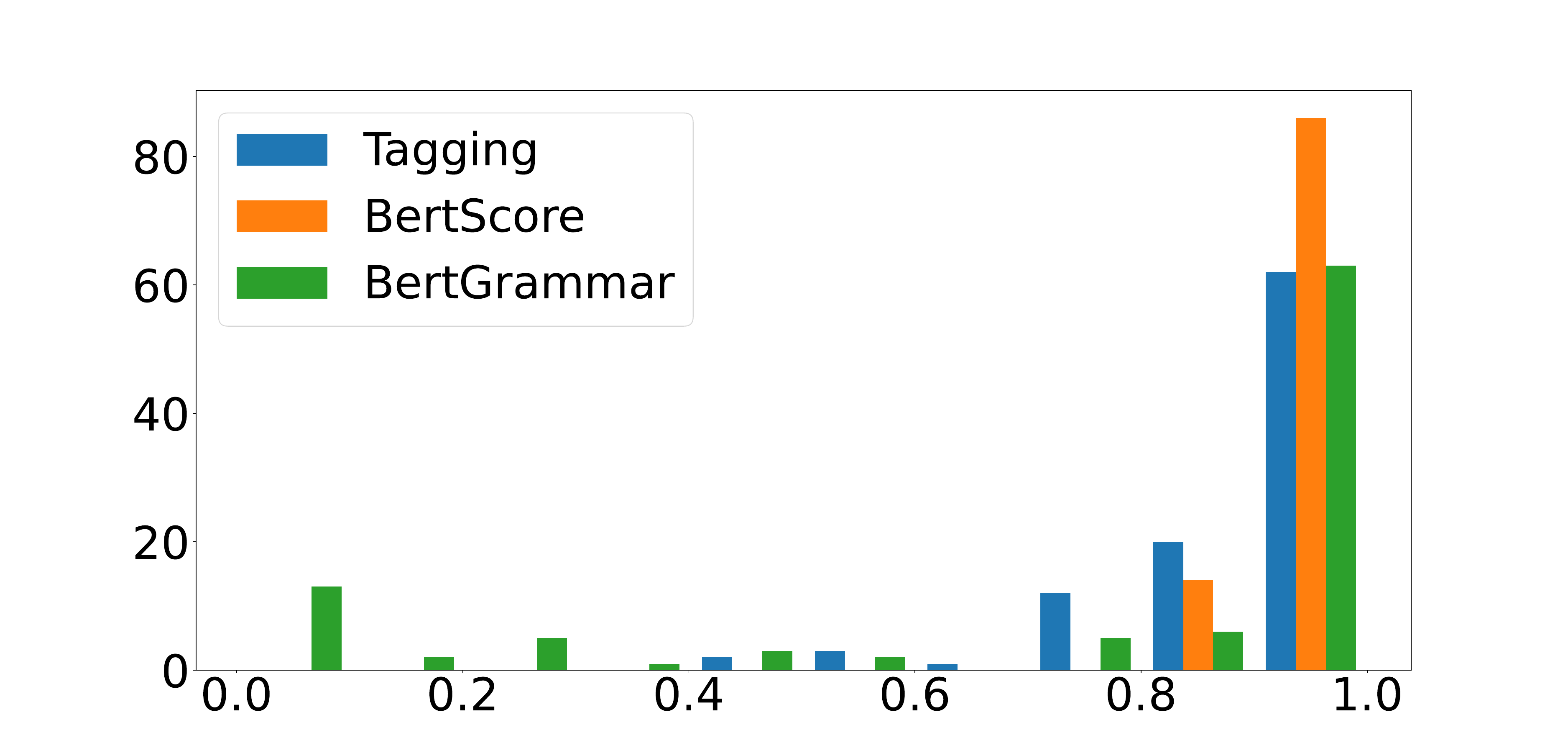}
    \caption{Humans}
\end{subfigure}
\begin{subfigure}{.45\textwidth}
    \centering
    \includegraphics[width=1.1\linewidth]{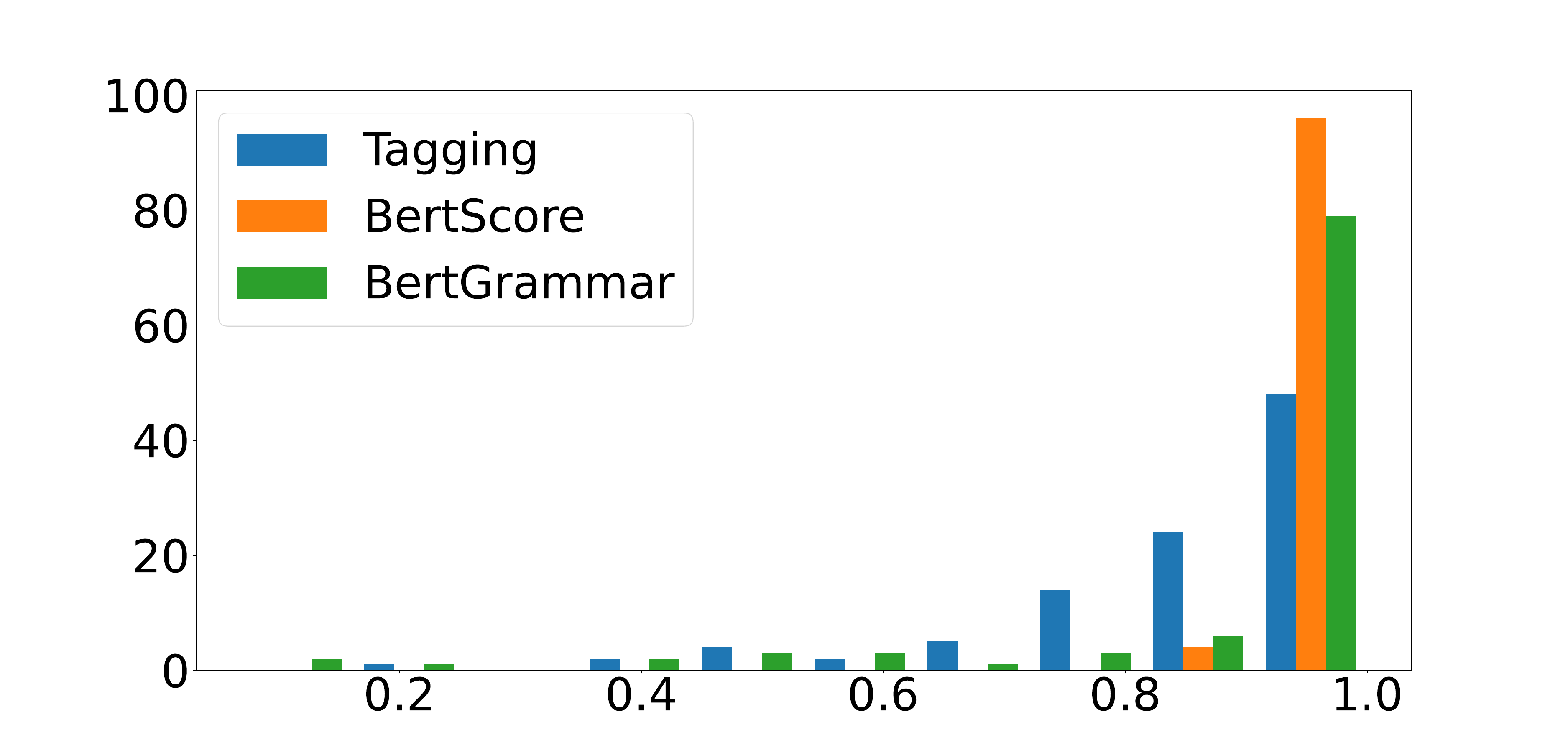}
    \caption{OFA}
\end{subfigure}
\\
\begin{subfigure}{.45\textwidth}
    \centering
    \includegraphics[width=1.1\linewidth]{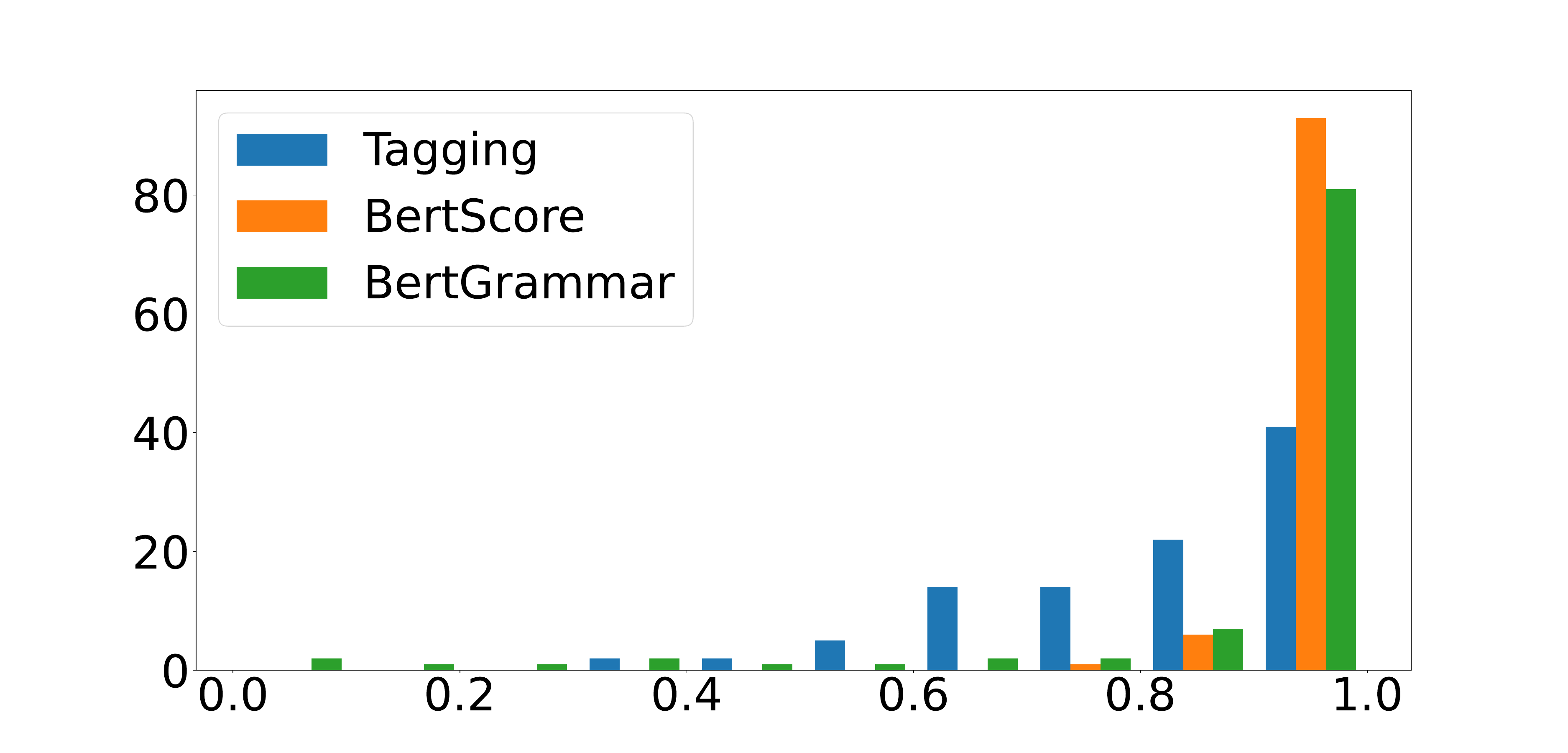}
    \caption{BLIP2}
\end{subfigure}
\begin{subfigure}{.45\textwidth}
    \centering
    \includegraphics[width=1.1\linewidth]{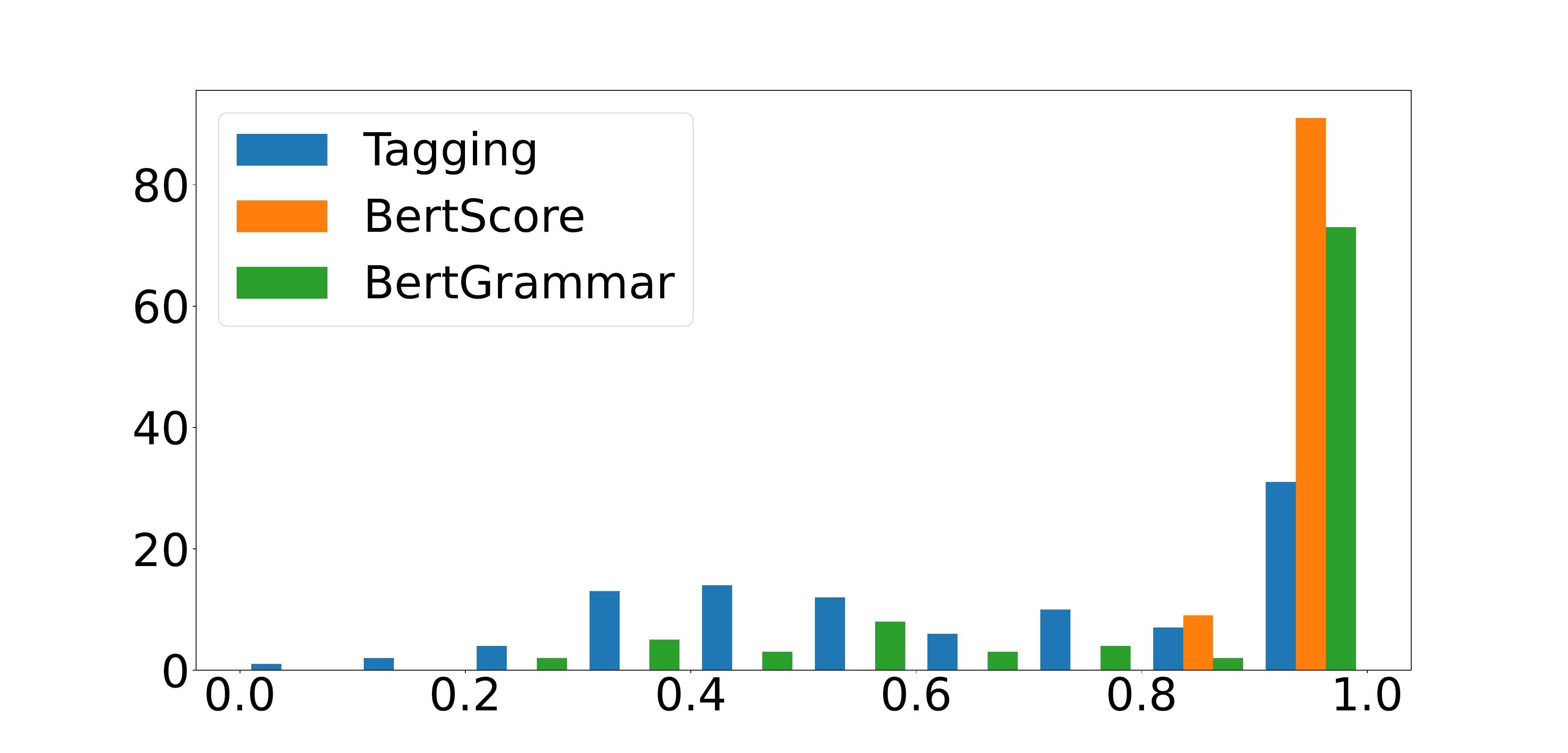}
    \caption{M$^2$}
\end{subfigure}
\\
\begin{subfigure}{.45\textwidth}
    \centering
    \includegraphics[width=1.1\linewidth]{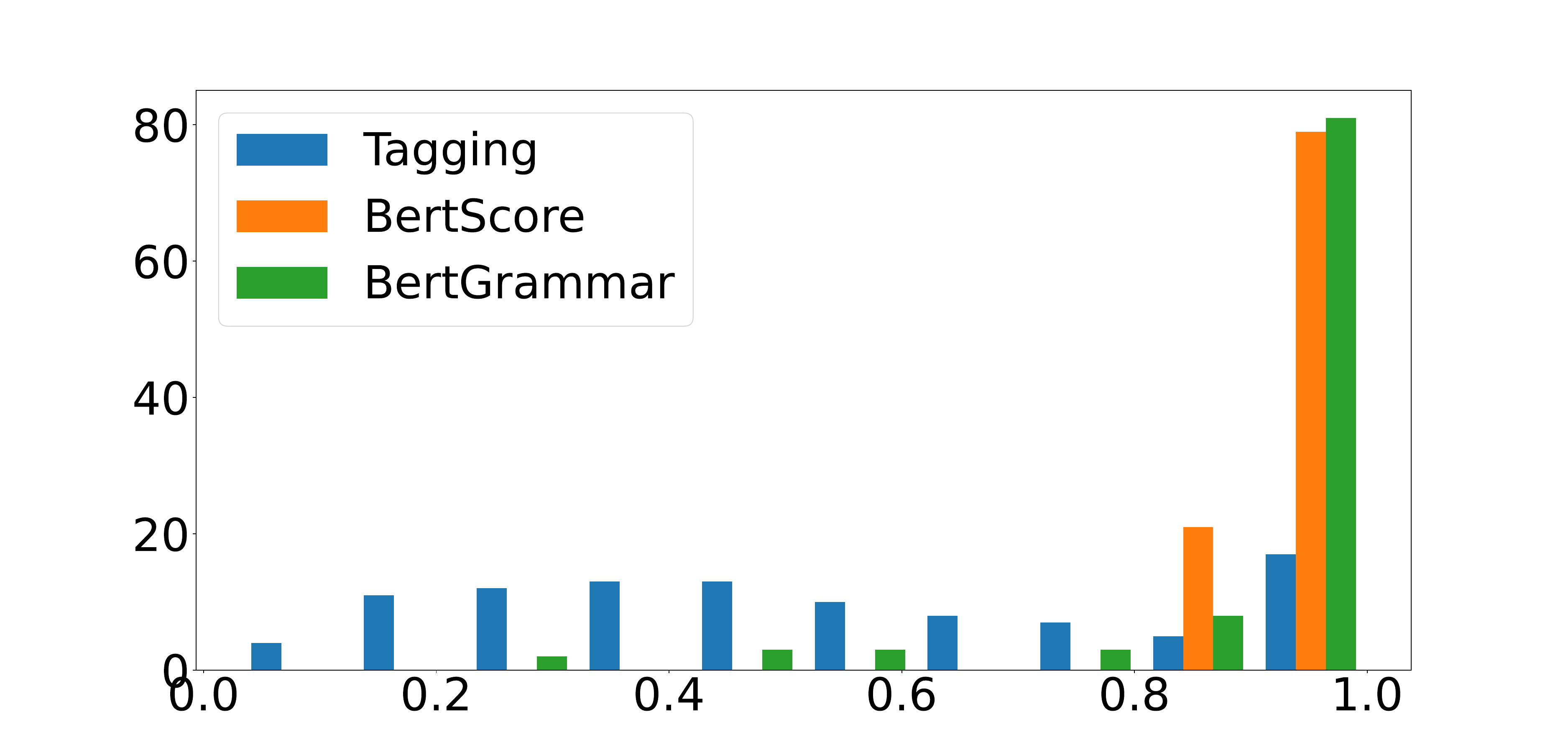}
    \caption{Convcap}
\end{subfigure}
\begin{subfigure}{.45\textwidth}
    \centering
    \includegraphics[width=1.1\linewidth]{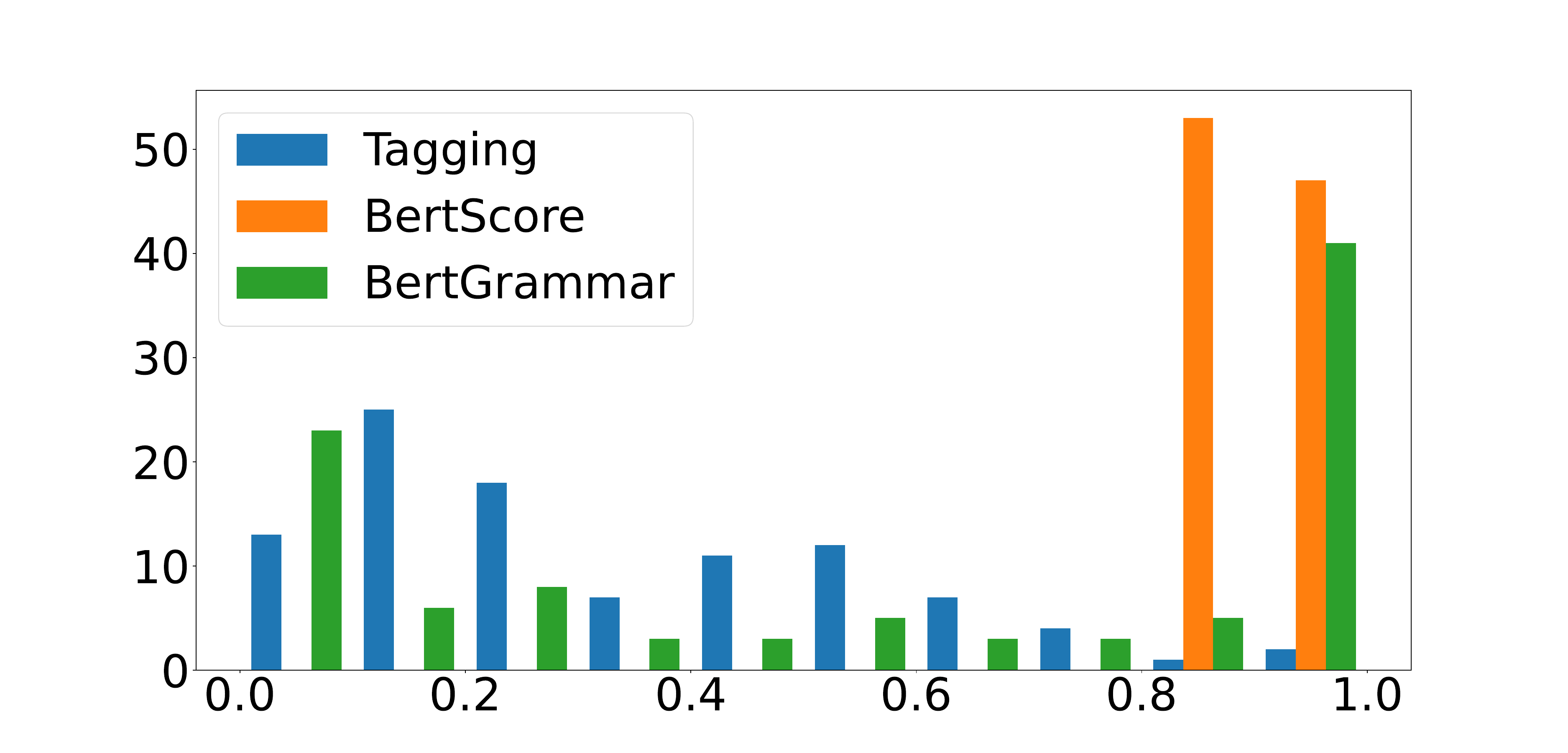}
    \caption{Show, Attend and Tell (SAAT)}
\end{subfigure}
\caption{Scores of the models with the human score (Tagging) compared to BertScore and BertGrammar }
\label{bertvsbergrammar}
\end{figure}

Next, a comparison between the BERT-based metrics in Figure~\ref{bertvsbergrammar} is shown. One can see that for Humans, OFA, and BLIP2, the automatic metrics give high scores to most of the captions, whereas human tagging tends to distribute the scores more widely. That might be an effect of the scale. In the case of Convcap, we can see that humans give distributed scores to the captions, which can not be seen with the other metrics. For the captions of the SAAT model, similar shapes can be seen between the human tagging and the BertGrammar model, suggesting that the penalization given by humans may have been due to grammatical errors.

\begin{figure}[!h]
\centering
\begin{subfigure}{.45\textwidth}
    \centering
    \includegraphics[width=1.1\linewidth]{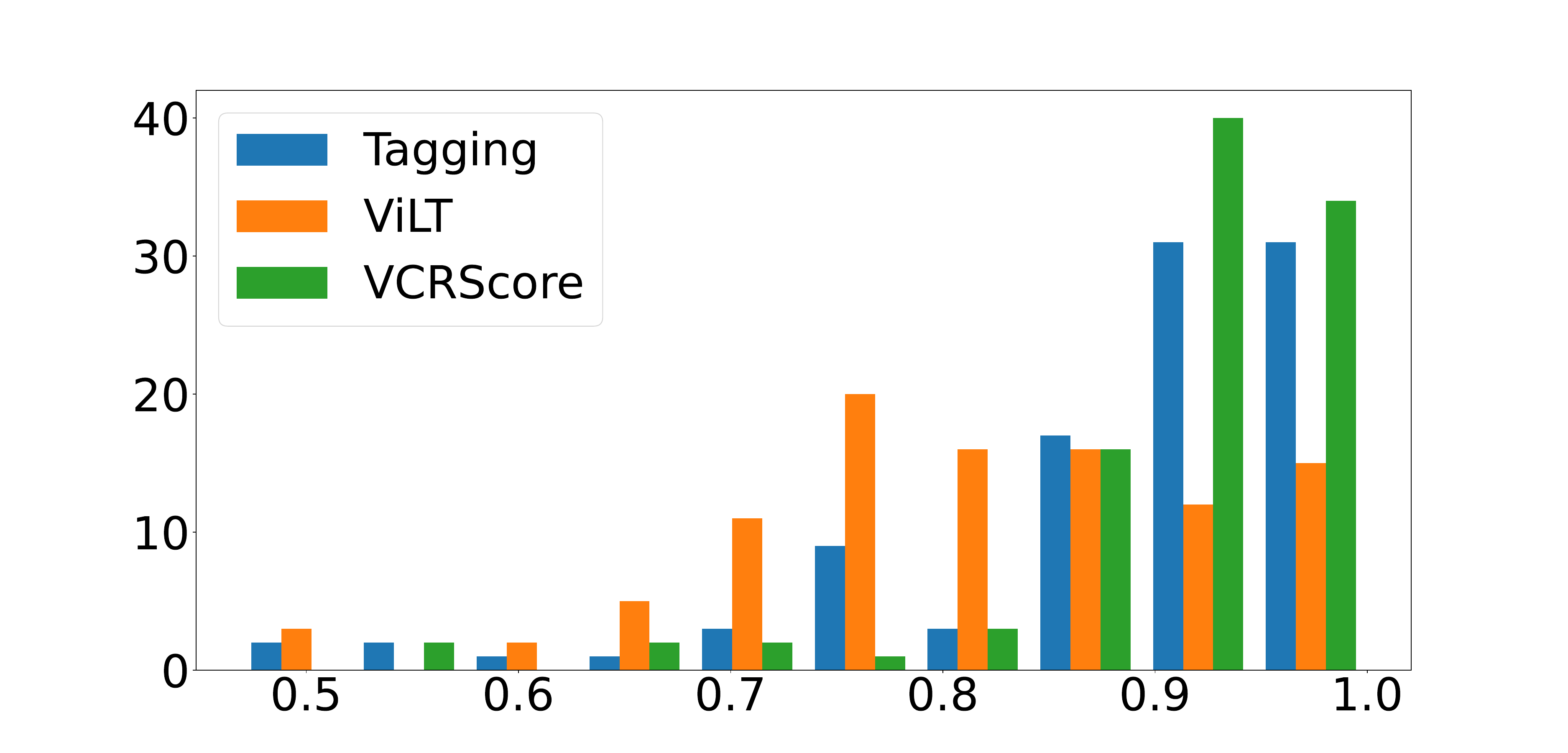}
    \caption{Humans}
\end{subfigure}
\begin{subfigure}{.45\textwidth}
    \centering
    \includegraphics[width=1.1\linewidth]{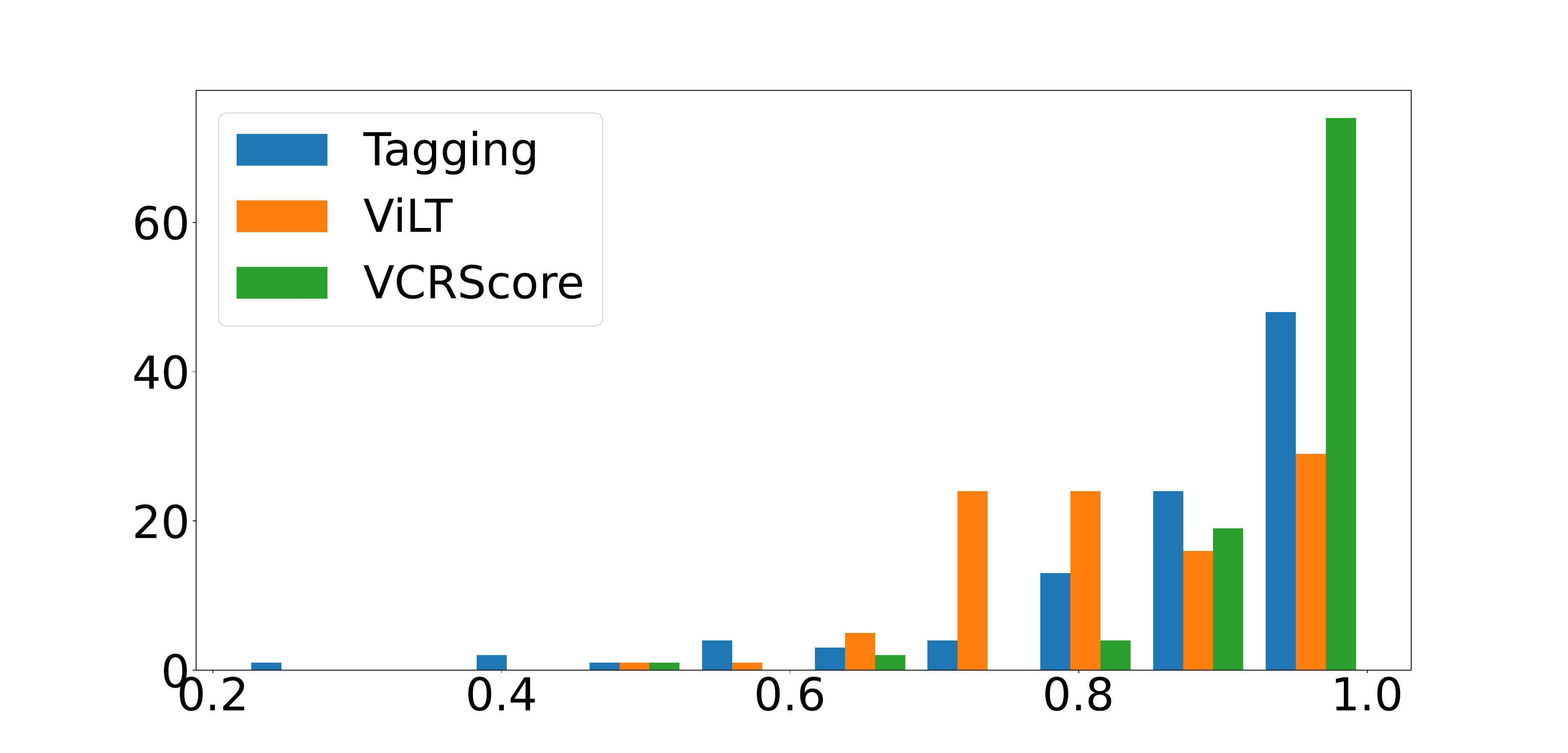}
    \caption{OFA}
\end{subfigure}
\\
\begin{subfigure}{.45\textwidth}
    \centering
    \includegraphics[width=1.1\linewidth]{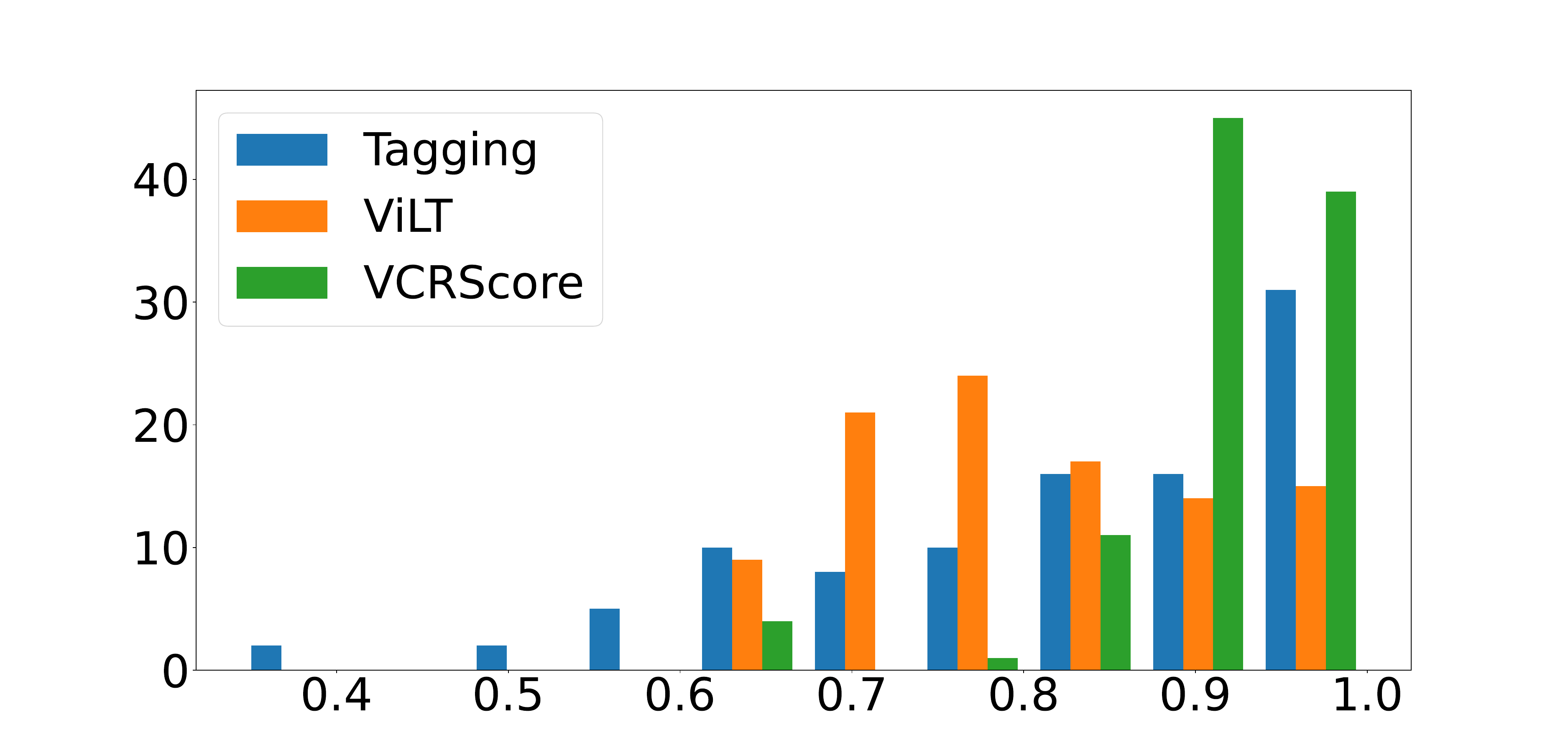}
    \caption{BLIP2}
\end{subfigure}
\begin{subfigure}{.45\textwidth}
    \centering
    \includegraphics[width=1.1\linewidth]{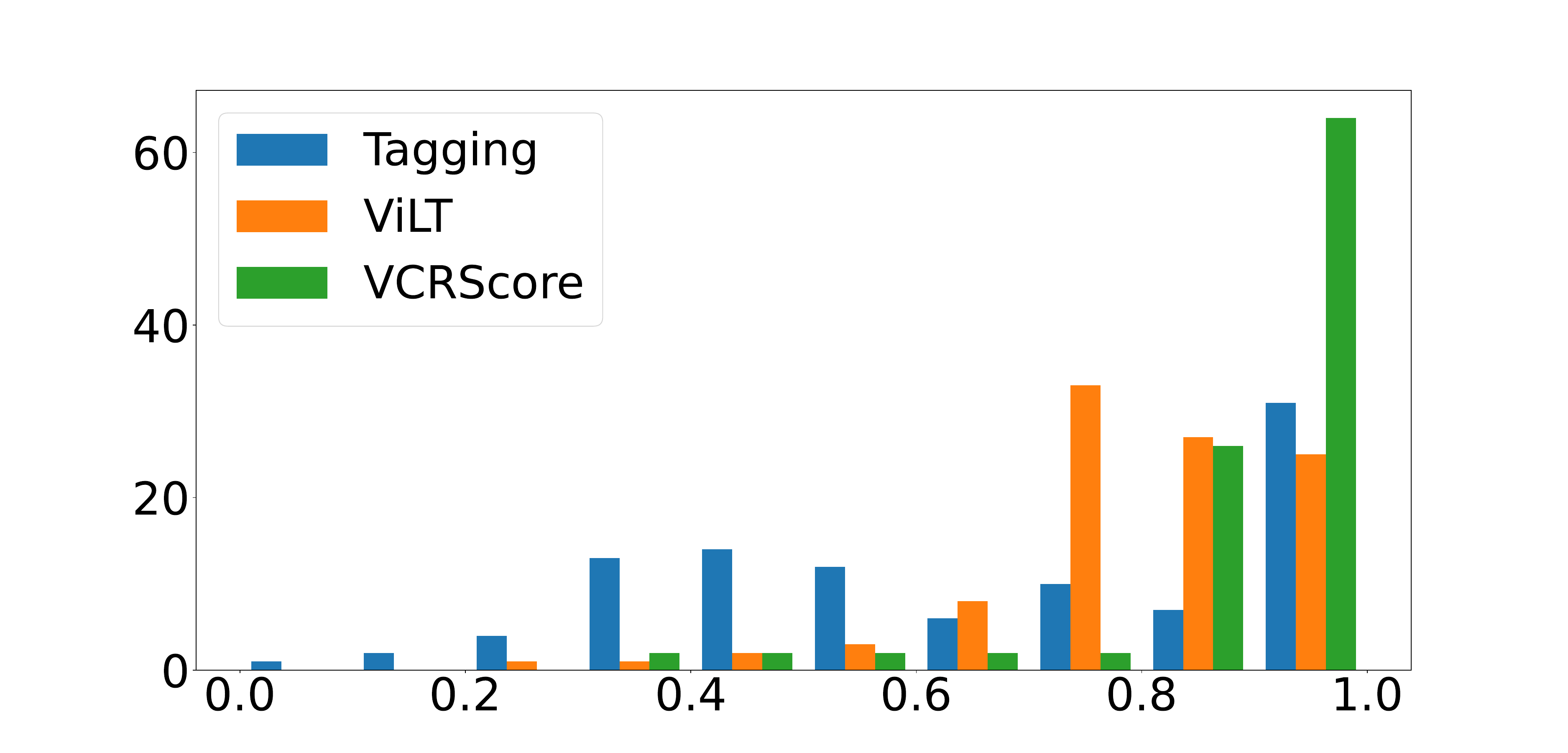}
    \caption{M$^2$}
\end{subfigure}
\\
\begin{subfigure}{.45\textwidth}
    \centering
    \includegraphics[width=1.1\linewidth]{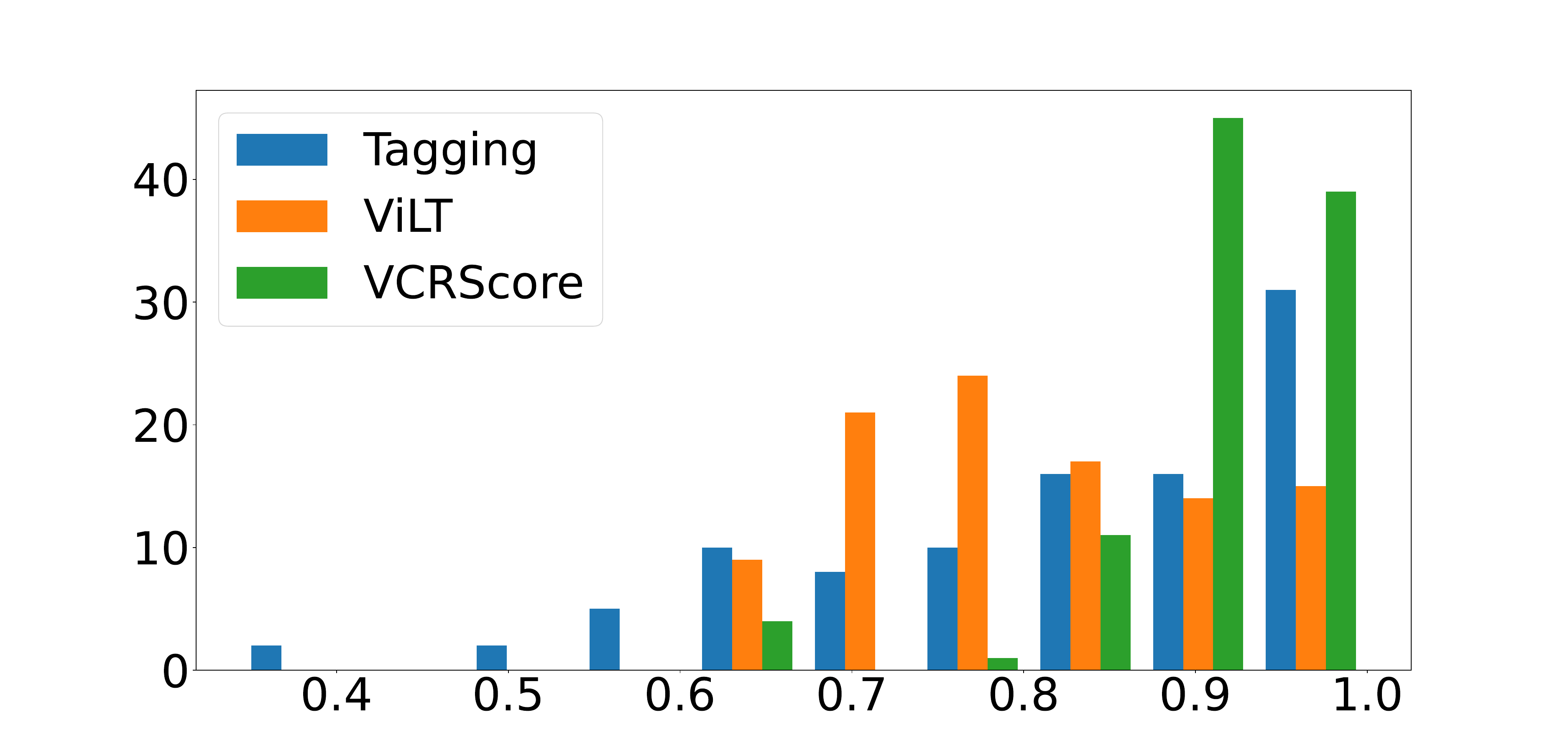}
    \caption{Convcap}
\end{subfigure}
\begin{subfigure}{.45\textwidth}
    \centering
    \includegraphics[width=1.1\linewidth]{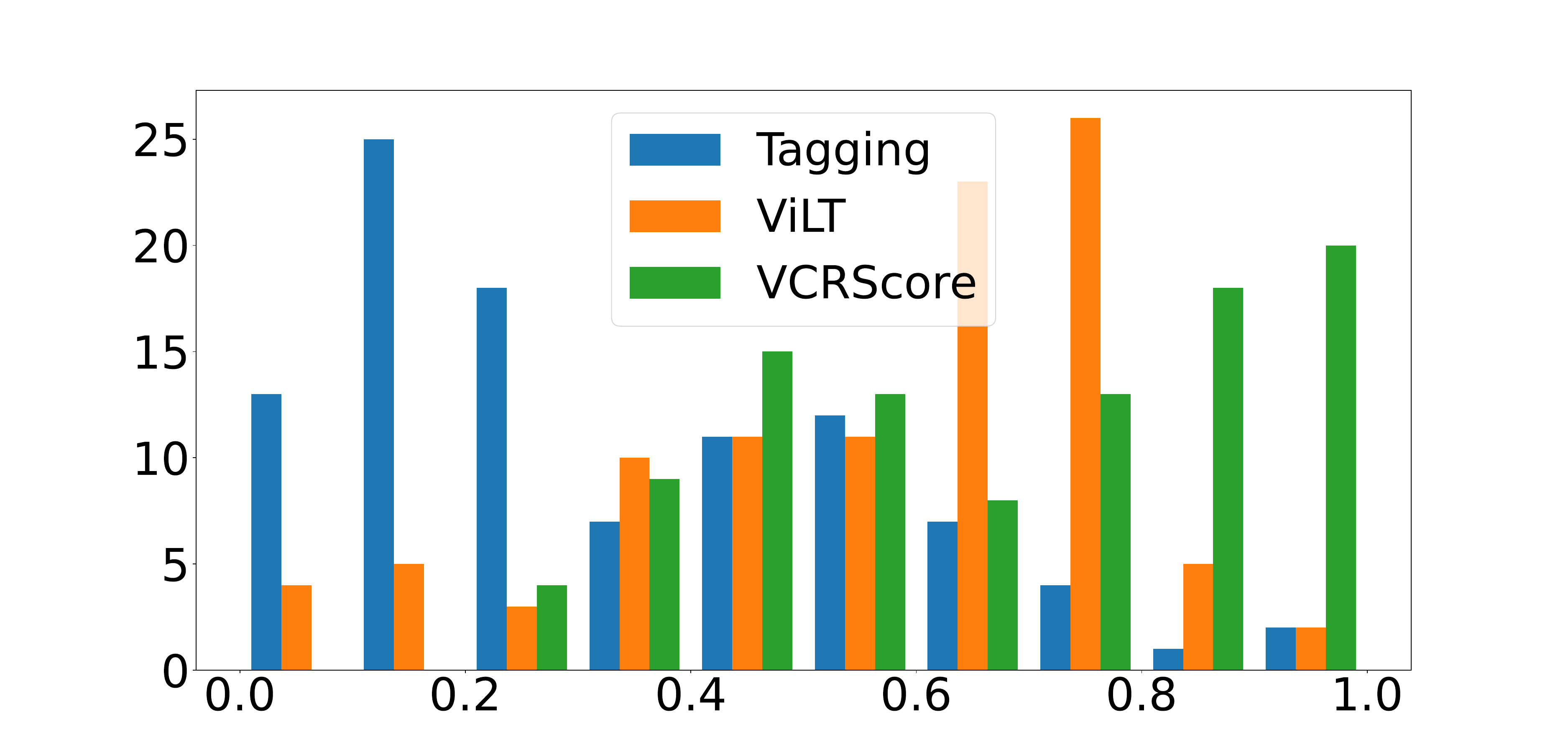}
    \caption{Show, Attend and Tell (SAAT)}
\end{subfigure}
\caption{Scores of the models with the human score (Tagging) compared to ViLT and VCRScore (our proposal).}
\label{viltvsVCRScore}
\end{figure}

In Figure~\ref{viltvsVCRScore},  our proposed VCRScore and the ViLT were compared since the former uses the latter. From the Human, OFA, and BLIP2 models, one can see that our proposal is more similar to the human tagging that the ViLT is. The captions from M$^2$ are evaluated differently by both automatic metrics compared to human tagging. This might be produced by the captions from M$^2$ ending abruptly and not being recognized by both automatic metrics. For the Convcap captions, one can see the VCRScore has a shape that is more similar to human tagging than ViLT.
Undoubtedly, there is no metric following the exact shape of the tagging histogram; sometimes, some metrics are more correlated in the middle of the histogram or at the end or beginning of it. However, definitively, the most similar histogram to human tagging is the one generated with the VCRScore proposal.

If one wants to analyze ranking positions instead of score values, an extra comparison was made, where each model was ranked according to their results expressed in each different evaluation metric. This means that according to, for instance, BLEU, each model is located in a position on the rank (from the best to the worst), and it does not matter if the difference between first and second place is very small or huge. We used as the correct ranking produced by the human tagging process, and all the rankings generated by each metric are compared to it in terms of
Spearman correlation.
This is, for the human tagging, the ranking was Humans, OFA, BLIP2, M$^2$, Convcap, and SAAT, in that order where the first is the best one and the worst the last one.

\begin{figure}[!h]
    \centering
    \includegraphics[width=.8\linewidth]{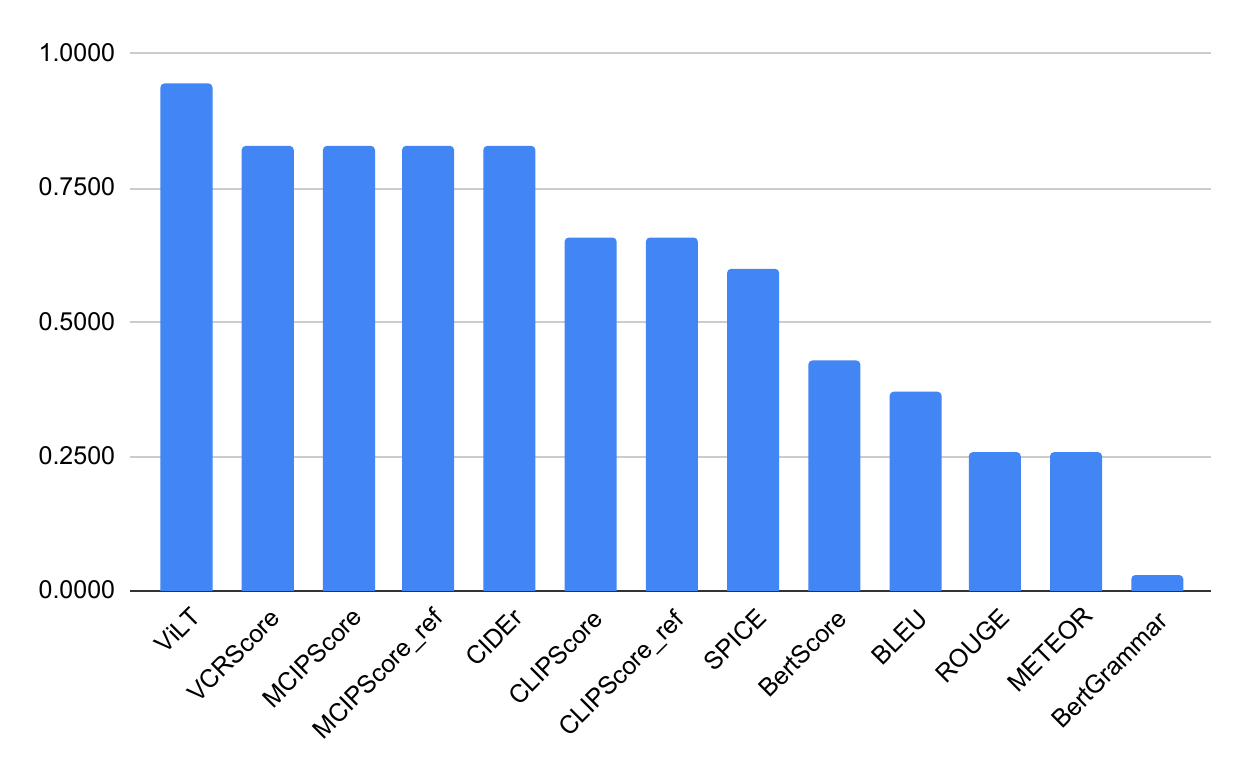}
    \caption{Correlations on positions between the metrics and the tagging process}
    \label{fig:corrpositions}
\end{figure}

Figure~\ref{fig:corrpositions} shows how each metric correlates to the ranking generated by the human taggers. Here, it can be seen how ViLT, VCRScore, MCIPScore, and MCIPScore\_ref produce the most similar ranking compared to the human-generated. The worst correlated are those related to a lexical approach and, surprisingly, BertGrammar.
Here, one can see that both positions and score values are advantageous in determining which model produced a better result than the other.

\begin{figure}[!h]
    \centering
    \includegraphics[width=1\linewidth]{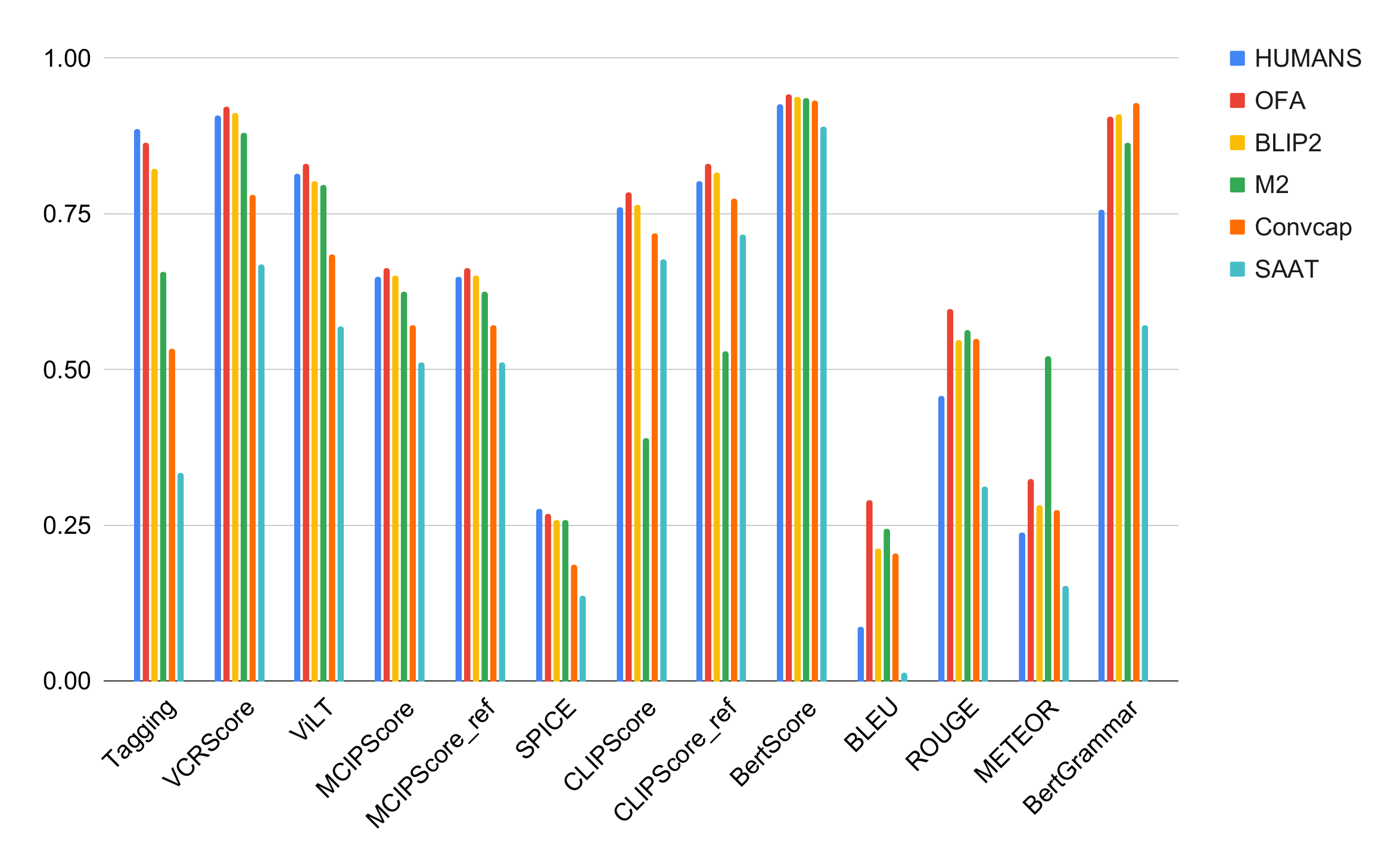}
    \caption{How each metric ranks all the models}
    \label{fig:corrpositions_2}
\end{figure}

As the final comparison, Figure~\ref{fig:corrpositions_2} shows how each metric ranks each model in rank and how high or low the score is. In the figure, all the performance metrics are located on the x-axis, and on the y-axis is the score they produce. Each color line reflects each model score on average; the first value presented is the human tagging; here it can be seen in the figure how they are ranked the models first place is humans (blue), second place is OFA (red), third place is BLIP2 (yellow), fourth place is M$^2$ (sky blue), fifth place is Convcap (green), and finally, last place is the SAAT model (orange).

Here, it can be seen that SPICE scores are very low in all the models. Although it was not a lower correlated rank, its scores for all the models are very low compared with the other metrics. BLEU and METEOR show similar behavior. In the contrary case, BertGrammar is lowest correlated in ranking positions but produces higher scores. BertScore is the most generous one because it assigned the highest scores to each model, all on average over 0.89. VCRScore and ViLT produce scores similar in ranking and magnitude to human tagging.

Through this experimental design, as well as results, we want to show how complex the performance evaluation is, whether by rank or score and how it is tough to be exactly the same as Human Tagging. Nevertheless, outstanding insights can be resumed from all these experiments.
First, metrics based on embedding or trained models are superior to those based on lexical aspects such as n-grams comparison. Second, not all the metrics provide the whole range of scores because most metrics are defined to give a value between $[0-1]$; maybe if one can stretch their values, more significant differences can be observed; this is because we analyzed and compared them in terms of positions.
Third, the usage of embeddings or models not currently used as performance metrics had a very outstanding performance; for instance, the adaptation of the version of CLIP for retrieval option into the ClipScore metric significantly improved the correlation with human tagging (MCIPScore and MCIPScore\_ref). The same happened with the ViLT model, which is not used to evaluate but achieved one of the most correlated performances.
Fourth, our generated metric VCRScore, which was trained to behave similarly to humans, did not reach the expected performance. Although it was the most correlated and best option regarding score and ranking, higher results were expected.
Fifth, humans generally evaluated the models with a higher range of values than the automatic metrics.

\section{Conclusions}
\label{sec:conclusions}
This paper proposed a new metric to evaluate the performance of the Image Captioning advancements.
To do that, we built a human-tagged dataset through a manual evaluation of four humans, their agreement, and how they evaluated a set of image captioning models from the state-of-the-art were analyzed.
In addition, a larger dataset was built through resources available on the internet for the community for both training and testing the proposals.

The VCRScore was proposed based on the new version of the CLIP proposal for information retrieval (MCIPScore and MCIPScore\_ref) as well as the vision and language transformer (ViLT). VCRScore is a metric trained to correlate with human tagging or evaluation.
A deep comparison through several experiments was made, using a set of well-known and used metrics for the state-of-the-art, from old to newer ones.

All the metrics were evaluated and compared with the human tagging, as well as how they ranked all the models tested and how big or small their score distributions were. Interesting findings were reached, and one can conclude that despite achieving a better solution, there are many challenges to deal with the better evaluation of the image captioning task.


\bibliographystyle{abbrv}
\bibliography{bibliography}

\end{document}